\documentclass[review]{article}

\usepackage{framed,multirow}

\usepackage{times}
\usepackage{bbold}
\usepackage{epsfig}
\usepackage{graphicx}
\usepackage{amsmath}
\usepackage{amsthm}
\usepackage{amssymb}
\usepackage{subcaption}
\usepackage{multirow}
\usepackage{lineno,hyperref}
\usepackage{graphicx}
\modulolinenumbers[5]
\newcommand{\argmin}{\mathop{\mathrm{argmin}}}
\newcommand*{\Scale}[2][4]{\scalebox{#1}{$#2$}}%
\usepackage{makecell}
\usepackage[dvipsnames]{xcolor}
\usepackage{booktabs}
\usepackage{arydshln}

\usepackage{latexsym}

\usepackage{url}
\usepackage{xcolor}
\usepackage{authblk}
\definecolor{newcolor}{rgb}{.8,.349,.1}

\title{A formal approach to good practices in Pseudo-Labeling for Unsupervised Domain Adaptive Re-Identification}

\author[1,2,*]{Fabian DUBOURVIEUX} 
\author[1]{Romaric AUDIGIER}
\author[1]{Angélique LOESCH}
\author[2]{Samia AINOUZ}
\author[2]{Stéphane CANU}
\date{}
\affil[1]{\textit{Université Paris-Saclay, CEA, List}, \textit{F-91120, Palaiseau, France}}
\affil[2]{\textit{Normandie Univ, INSA Rouen, LITIS},
Av. de l'Université le Madrillet 76801 Saint Etienne du Rouvray, France}
\affil[*]{Corresponding author: Fabian DUBOURVIEUX, fabian.dubourvieux@cea.fr}

\begin{document}

\maketitle



\begin{abstract}
The use of pseudo-labels prevails in order to tackle Unsupervised Domain Adaptive (UDA) Re-Identification (re-ID) with the best performance. Indeed, this family of approaches has given rise to several UDA re-ID specific frameworks, which are effective.
In these works, research directions to improve Pseudo-Labeling UDA re-ID performance are varied and mostly based on intuition and experiments: refining pseudo-labels, reducing the impact of errors in pseudo-labels... It can be hard to deduce from them general good practices, which can be implemented in any Pseudo-Labeling method, to consistently improve its performance. To address this key question, a new theoretical view on Pseudo-Labeling UDA re-ID is proposed. The contributions are threefold: (i) A novel theoretical framework for Pseudo-Labeling UDA re-ID, formalized through a new general learning upper-bound on the UDA re-ID performance. (ii) General good practices for Pseudo-Labeling, directly deduced from the interpretation of the proposed theoretical framework, in order to improve the target re-ID performance. (iii) Extensive experiments on challenging person and vehicle cross-dataset re-ID tasks, showing consistent performance improvements for various state-of-the-art methods and various proposed implementations of good practices.
\end{abstract}


\section{Introduction}

Recent advances on Supervised Re-Identification (re-ID) have led to excellent performance on widely used public re-ID datasets (\cite{ye2021deep,ni2021adaptive}). However, performance suffers from a significant drop when re-ID models are tested cross-dataset, i.e., on images of a target context different from the training context. In this practical setting, to avoid a costly manual annotation of new target datasets, Unsupervised Domain Adaptation (UDA) has gained interest among machine learning and computer vision researchers. The UDA goal is to transfer valuable knowledge of a re-ID model from a source domain to a target domain, without any additional identity (ID) label on the target domain. Unsupervised Domain Adaptive re-ID (UDA re-ID) methods have been designed to tackle the open-set nature of re-ID in which classes of individuals at test time are different from those seen during the training stage. A first line of works in UDA re-ID investigated \emph{Domain Translation}. This family of approaches aims at bridging the gap between the source and target domains in the image space with style-transfer generative models (\cite{wei2018person, deng2018image, peng2019cross, zhong2018generalizing, qi2019novel}) or by constraining the feature space to learn a domain-invariant discriminative representation (\cite{chang2019disjoint, li2018adaptation, li2019cross, lin2018multi, mekhazni2020unsupervised}). Nevertheless, Domain Translation based approaches fail to learn target-specific discriminative features, thus limiting the re-ID performance achieved on the target domain.\\
Recent UDA re-ID approaches widely rely on the use of \emph{pseudo-labels} for the target domain data (\cite{song2020unsupervised, zhang2019self, jin2020global, tang2019unsupervised, zhai2020ad, zou2020joint, yang2019asymmetric, chendeep, ge2020mutual, zhai2020multiple, zhao2020unsupervised, zou2020joint,peng2020unsupervised, Zhang_2021_CVPR, Yang_2021_CVPR}). In fact, learning with these target pseudo-labeled samples can lead to a better identity-discriminative representation on the target domain. For this purpose, researchers designed a wide range of Pseudo-Labeling frameworks, based on a wide variety of modeling choices. Therefore, various directions have been explored to improve pseudo-label based UDA re-ID approaches: some works focus on improving the predicted pseudo-labels (\cite{jiattention,zhai2020ad, tang2019unsupervised, zou2020joint, Chen_2021_CVPR, Zheng_2021_CVPR, zheng2021online, Zhang_2021_CVPR}), others on reducing the impact of pseudo-label errors during training (\cite{ge2020mutual, zhao2020unsupervised, zhai2020multiple, jin2020global, liu2020domain, fu2019self, Xuan_2021_CVPR,lin2020unsupervised, Yang_2021_CVPR, lin2020unsupervised, Yang_2021_CVPR, feng2021complementary, ge2020self}). These Pseudo-Labeling practices, integrated in specific Pseudo-Labeling methods, have been shown experimentally to improve the performance. However, there is no general theoretical framework that can justify these practices, highlight the conditions for their effectiveness and explain how they improve the performance .\\
Moreover, some of these practices are not unanimously agreed among Pseudo-Labeling UDA re-ID work: the most common example is leveraging (\cite{dub2020, ge2020self, Isobe_2021_ICCV, dubourvieux2021improving}) or alternatively discarding (\cite{zhang2019self, zhai2020ad}) the available ground-truth labeled source data to optimize the model jointly with the pseudo-labeled target data. 
Note that, the majority of Pseudo-Labeling UDA re-ID approaches do not use the available source training set (\cite{zhang2019self, zhai2020ad, song2020unsupervised, jin2020global, tang2019unsupervised, zhai2020ad, zou2020joint, yang2019asymmetric, chendeep}) after having access to target pseudo-labels. However, some recent works (\cite{dub2020, ge2020self, dubourvieux2021improving, Isobe_2021_ICCV}) show experimentally that the source data, continuously used in addition to the target pseudo-labeled data, can improve re-ID performance on the target domain. Can the source data help to improve the cross-dataset re-ID performance of any Pseudo-Labeling approach ? Are there some conditions under which any Pseudo-Labeling UDA re-ID method can benefit from the source data, after having access to target pseudo-labels? These questions are also left unanswered.\\
We believe the lack of general insight on these key UDA re-ID questions is due to the predominance of empirical-oriented works for UDA re-ID. That’s why this work seeks to answer these questions with a general theoretical framework, and to highlight good practices to design an effective Pseudo-Labeling UDA re-ID method. 
To this end, this work proposes three contributions that can be summarized as follows:
\begin{enumerate}
\item Theoretical development to find a new upper-bound
\item General good practices for Pseudo-Labeling UDA re-ID approaches, deduced from the theory, and therefore providing a better insight of their role in improving the cross-domain re-ID performance
\item Experiments with different proposed implementations of these good practices, showing performance improvements of recent state-of-the-art methods by implementing the whole proposed good practices, on various person and vehicle cross-domain benchmarks
\end{enumerate}

\begin{figure*}[t!]
\includegraphics[width=\linewidth]{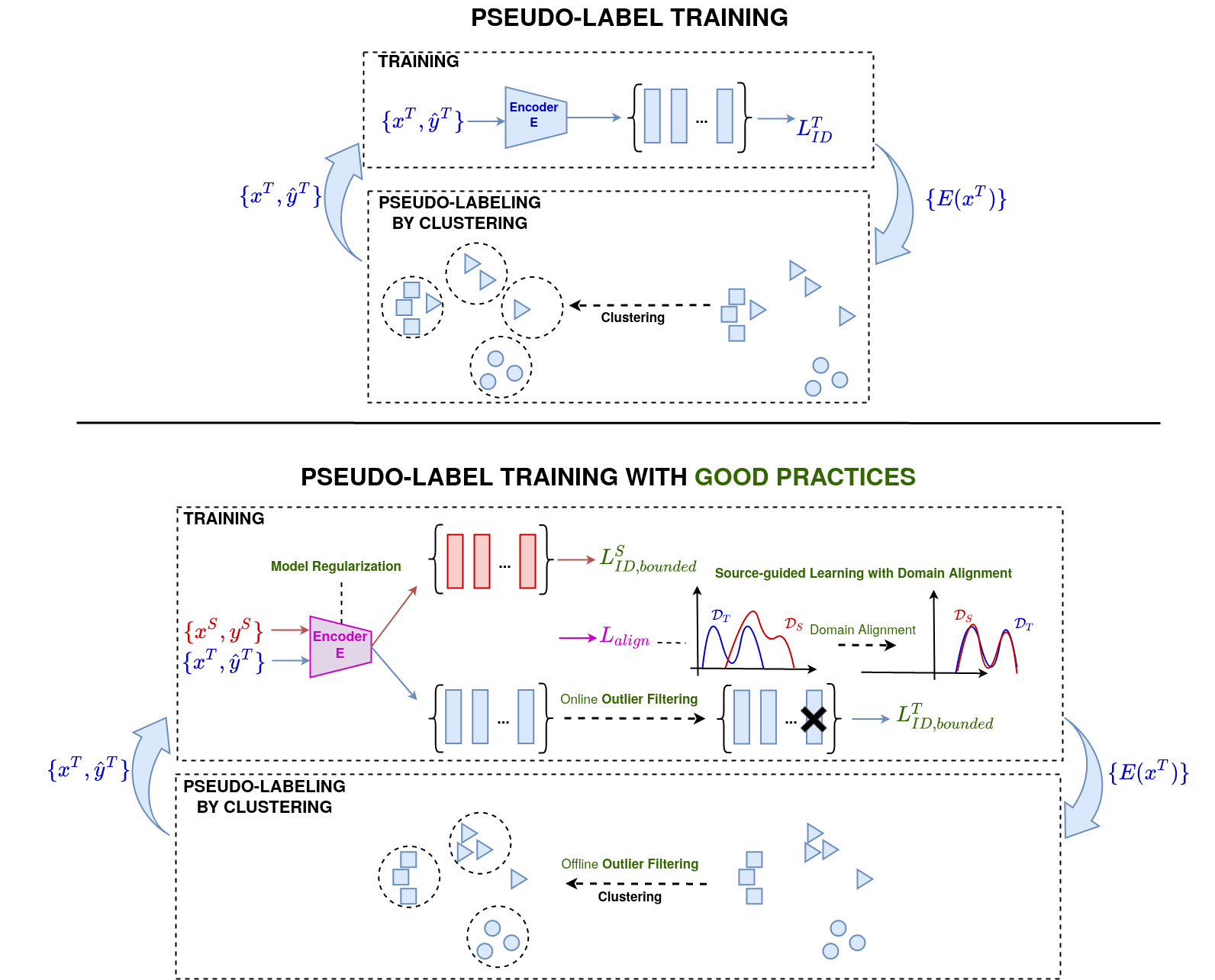}
\caption{Classical Pseudo-Label Training is illustrated at the top of the figure. At the bottom, the Pseudo-Label Training with good practices is illustrated. Good practices are derived from analysis of our new learning bound for Pseudo-Labeling, to improve UDA re-ID performance. These good practices, when followed and implemented in Pseudo-Labeling method, aim at improving the re-ID performance on the target domain.
These good practices, represented in green on the figure, are: \emph{Source-guided Learning with Domain Alignment}, using a \emph{Bounded Loss} for re-ID learning, \emph{Model Regularization} and \emph{Outlier Filtering} (performed offline and/or online). $L^S_{ID,bounded}$ and $L^T_{ID,bounded}$ are bounded loss functions resp. defined for the source ($\{x^S,y^S\}$) and target ($\{x^T,\hat{y}^T\}$) samples to learn re-ID features. $L_{align}$ is a loss that penalizes a domain discrepancy measure between the source and the target distributions (resp. $\mathcal{D}_S$ and $\mathcal{D}_T$)  in the similarity feature space.}
    \label{fig:good_practice}
\end{figure*}

The paper is structured as follows: in Sec.~\ref{sec:related}, the relevant existing work is reviewed. In Sec.~\ref{sec:theory}, notations, concepts and models on which our theoretical framework is based are introduced. Then in Sec.~\ref{sec:bound}, a new upper-bound on the target re-ID performance is built, taking into account the errors in the pseudo-labels, as well as the exploitation of the source data. In Sec.~\ref{sec:theory_to_practice}, insight and a set of general good practices for improving the UDA re-ID performance are deduced from the theoretical bound. After introducing the experimental setting and proposed implementations for the good practices in Sec.~\ref{sec:experiments}, extensive experiments are conducted in Sec.~\ref{sec:results} on person and vehicle UDA re-ID benchmarks, and are discussed to validate the performance improvement obtained by following the good practices in a Pseudo-Labeling method.

\section{Related Work}
\label{sec:related}

\subsection{Pseudo-Labeling methods for UDA re-ID.}
\label{sec:state-of-the-art_udareid}

Pseudo-Labeling approaches have been the main research direction for UDA re-ID due to their better performance on the target domain compared to UDA re-ID based on Domain translation (i.e. style transfer (\cite{wei2018person, deng2018image, peng2019cross, zhong2018generalizing, qi2019novel}) or domain-shared feature space (\cite{chang2019disjoint, li2018adaptation, li2019cross, lin2018multi, mekhazni2020unsupervised}) techniques). These pseudo-label based methods seek to learn a target domain identity-discriminative representation through supervision by using artificial/generated labels for the target samples. These methods generally leverage a source-trained re-ID model to initialize pseudo-identity labels for target data by feature representation clustering. Most Pseudo-Labeling methods (\cite{song2020unsupervised, zhang2019self, jin2020global, tang2019unsupervised, zhai2020ad, zou2020joint, yang2019asymmetric, chendeep, ge2020mutual, zhai2020multiple, zhao2020unsupervised, zou2020joint,peng2020unsupervised, yu2019unsupervised, zhong19enc, luogeneralizing}) are built on a self-learning iterative paradigm which alternates between (i) optimization of the model to learn target re-ID feature in a supervised way with the pseudo-labeled target images (ii) pseudo-label refinement by feature clustering (see the top of Fig.~\ref{fig:good_practice} for an illustration of the classical Pseudo-Labeling self-learning cycle).\\
Pseudo-Labeling works explore various research directions to improve the target re-ID performance. The main research direction for Pseudo-Labeling approaches focuses on limiting the impact of pseudo-label errors on the final UDA re-ID performance. To this end, various Pseudo-Labeling methods have been designed, for example by using teacher-student or ensemble of expert models (\cite{ge2020mutual, zhao2020unsupervised, zhai2020multiple}), by optimizing on global distance distributions (\cite{jin2020global, liu2020domain}), or by leveraging local features (\cite{fu2019self}, intra-inter camera features (\cite{Xuan_2021_CVPR,lin2020unsupervised, Yang_2021_CVPR}), multiple cluster views (\cite{feng2021complementary}), or even by leveraging class-centroid and instance feature memory banks with contrastive learning (\cite{ge2020self}). Another line of work focuses on pseudo-label refinement, by using attention-based models (\cite{jiattention}), by combining pseudo-labels with domain-translation/generative methods (\cite{zhai2020ad, tang2019unsupervised, zou2020joint, Chen_2021_CVPR}), by using online pseudo-labels (\cite{Zheng_2021_CVPR, zheng2021online}), label propagation (\cite{Zhang_2021_CVPR}) or by selecting better clustering hyperparameters (\cite{dubourvieux2021improving}). Other approaches focus on designing efficient sample selection and outlier detection strategies (\cite{yang2019asymmetric, chendeep}). Then, a recent line of work seeks to leverage the source knowledge during pseudo-label training (\cite{dub2020, ge2020self, Isobe_2021_ICCV, dubourvieux2021improving}), contrary to other work that discards the source data after pseudo-label generation for the target data.\\
While existing works design various Pseudo-Labeling approaches, with an empirical focus on the UDA re-ID performance of a specific method, ours proposes a general and theoretical view of Pseudo-Labeling UDA re-ID, focusing on better understanding of good practices applicable to any method. To our knowledge, the exception is the paper of UDAP (\cite{song2020unsupervised}) which offers a theoretical framework on Pseudo-Labeling UDA re-ID. However, our theoretical framework differs from UDAP on various aspects. First, our theoretical framework models directly the errors in the pseudo-labels contrary to UDAP. Moreover, we propose a Pseudo-Labeling learning upper-bound directly on the target re-ID performance, while UDAP focuses on a measure of clusterability on the target domain. In other words, the theoretical framework of UDAP aims at justifying the Pseudo-Labeling self-learning paradigm, while ours seeks to encompass all the lines of researches and good practices that improve empirically the UDA re-ID performance.


\subsection{Theoretical works on Pseudo-Labeling UDA classification.}

It is relevant to ask whether theoretical work exists for Pseudo-Labeling UDA classification, a more investigated task than re-identification in the machine learning field. However, to our knowledge, there is no work that jointly modeled the impact of using source data in addition to pseudo-labeled target data. One of the pioneering works on Pseudo-Labeling for UDA (\cite{shu2018dirt}), considers in its theoretical developments the use of pseudo-labels without the source data to tackle the case of ''non-conservative" domain adaptation. Their theoretical framework does not model the impact of errors in pseudo-labels on the classification accuracy nor the use of the source samples after pseudo-label generation. The closest theoretical work would be that of Ben David et al. (\cite{ben2010theory}), in which they consider the problem of finding a model that minimizes the risk on the target domain, by minimizing an empirical risk jointly with labeled source and labeled target data. Nevertheless, this Source-guided theoretical work is not completely applicable in our case, as it considers the use of ground-truth target labels rather than pseudo-labels.
From this point of view, the work of Natarajan et al. (\cite{natarajan2013learning}) models the impact of errors in the labels on the classification accuracy. However, the use of labeled data from another domain is not taken into account and therefore it is incomplete to model a domain adaptation problem. \\
This work proposes a Source-guided Pseudo-Labeling theoretical framework bridging the work of Ben David et al. (\cite{ben2010theory}) and Natarajan et al. (\cite{natarajan2013learning}), specifically thought for UDA re-ID, in order to understand with a theoretical and general view all existing modeling practices in the UDA re-ID literature. Contrary to existing work, ours aims at deducing general good practices for UDA re-ID from theoretical analysis and interpretation.


\section{Notations and concepts to establish a new learning bound for Pseudo-Labeling UDA re-ID}
\label{sec:theory}

To have a general and theoretical framework that tries to encompass the variety of Pseudo-Labeling UDA re-ID practices, we should model the use of the source labeled data in addition to the pseudo-labeled target data for training the re-ID model. To do so, we choose to establish a new learning upper-bound on the target re-ID performance, measured by an expected risk on the target domain. We expect this upper-bound to highlight the use of the source data and the pseudo-labeled target data during the training.\\
The first step will be to model and define this target expected risk for the re-ID task in Sec.~\ref{sec:theory_verif}. Then, we will focus on how to measure the impact of using the source data on the target re-ID performance, by defining different Domain Discrepancy measures in Sec.~\ref{sec:theory_uda}. Finally, we will model the use of pseudo-label for the target data in Sec.~\ref{sec:pseudo_label}. All these preliminary modeling put together will be used to establish the desired learning upper-bound in Sec.~\ref{sec:bound}.

\subsection{Definitions and notations for the re-ID problem.}
\label{sec:theory_verif}
In this paper, we are particularly interested in the re-ID problem. As re-ID can be learned by binary classification of pairs of images, as having the same identity or not (i.e. a verification task), we choose to formulate the re-ID problem this way. Indeed, reformulating the re-ID problem as a verification problem allows us to model it as a closed-set classification task. It is therefore expected that this modeling will simplify theoretical development by allowing us to reuse some results already established in other works for binary classification (\cite{ben2010theory}). 
Consequently, we consider an input space describing pairs of images $\mathcal{X} \in \mathbb{R}^p \times \mathbb{R}^p, p \in \mathbb{N}$ and an output space $\{-1;1\}$ where ``1'' represents the label assigned to a pair of images with the same identity (``-1" otherwise). Therefore, in this paper, $x \in \mathcal{X}$ will represent a pair of images (or a pair of image feature vectors). \\ 
To measure the re-ID performance, we need to define a loss function $\mathcal{L}$. Usually, the binary classification task is evaluated by the ``0-1" loss. However, as we would like to highlight afterward the influence of the loss bounds on the performance of our model, we choose a slight modification of this loss that we call the ``0-M" loss. Contrary to the ``0-1" loss, we expect with the ``0-M" loss to explicitly highlight the loss bound $M > 0$ in the established learning bound. Indeed, the loss bound may give some insight for Pseudo-Labeling, and using the ``0-1" loss, bounded by ``1", may hide this relevant information in a multiplicative interaction $\times 1$ with another term of the learning upper-bound.
The ``0-M" loss is defined by:
\begin{equation} 
\label{eq:0MLoss}
\forall y,y' \in \{-1;1\}, \mathcal{L}(y,y') =  \frac{M}{2} \|y-y' \| = M\mathbb{1}_{y \neq y'},
\end{equation} 
where $M > 0$ represents the loss bound since: $\forall y,y' \in \{-1;1\}, \| \mathcal{L}(y,y') \| \leq M$.

\noindent This work particularly focuses on the UDA re-ID task that we specify hereafter.

\subsection{Measuring the Domain Discrepancy for Unsupervised Domain Adaptation.}
\label{sec:theory_uda}

\subsubsection{General definitions and notations for UDA.}

 To model UDA, we consider two domains $S$ and $T$, resp. called the source and target domains. These domains can be described as pairs of distributions $S = (\mathcal{D}^S,f_S)$ and $T = (\mathcal{D}^T,f_T)$, where $\mathcal{D}^S$, $\mathcal{D}^T$ are resp. the source and target domain marginal input distributions defined on $\mathcal{X}$, and $f_S : \mathcal{X} \rightarrow \{-1;1\}$, $f_T : \mathcal{X} \rightarrow \{-1;1\}$ represent resp. the source and target domain (ground-truth) labeling functions.\\
UDA aims at finding a hypothesis function (also called a classifier) $h \in \mathcal{H} \subseteq \{-1;1\}^\mathcal{X}$ which minimizes the target expected risk $\epsilon_{T}(h,f_T)$ related to the loss function $\mathcal{L}$: $\epsilon_{T}(h,f_T) = \mathbb{E}_{x \sim \mathcal{D}^T}[ \mathcal{L}(h(x),f_T(x))]$. $\mathcal{L}$ being our defined ``0-M" loss (Eq.~\ref{eq:0MLoss}), the target expected risk $\epsilon_{T}(h,f_T)$ can more particularly be expressed as:
\begin{equation}
\begin{aligned}
\label{eq:expected_risk}
\forall h \in \mathcal{H}, \epsilon_{T}(h,f_T) &= \mathbb{E}_{x \sim \mathcal{D}^T}[ \mathcal{L}(h(x),f_{T}(x))]\\
&= \mathbb{E}_{x \sim \mathcal{D}^T}[ M\mathbb{1}_{h(x) \neq f_T(x)}].
\end{aligned}
\end{equation}
More generally, we also define $\forall h,h' \in \mathcal{H}, \epsilon_{T}(h,h') = \mathbb{E}_{x \sim \mathcal{D}^T}[ \mathcal{L}(h(x),h'(x))]$ as well as the notation shortcut $\forall h \in \mathcal{H}, \epsilon_{T}(h) = \epsilon_{T}(h,f_T)$. In the same way, we define the source expected risk $\forall h \in \mathcal{H}, \epsilon_{S}(h) = \epsilon_{S}(h,f_S) = \mathbb{E}_{x \sim \mathcal{D}^S}[ \mathcal{L}(h(x),f_{S}(x))] = \mathbb{E}_{x \sim \mathcal{D}^S}[ M\mathbb{1}_{h(x) \neq f_S(x)}]$ and $\forall h,h' \in \mathcal{H}, \epsilon_{S}(h,h') = \mathbb{E}_{x \sim \mathcal{D}^S}[ \mathcal{L}(h(x),h'(x))]$.\\
Under the UDA setting, we want to minimize $\epsilon_{T}$ given by Eq.~\ref{eq:expected_risk} w.r.t $h \in \mathcal{H}$, and for this, we have access to a set of i.i.d labeled source samples and a set of i.i.d unlabeled target samples.\\

\subsubsection{Measuring the domain gap.}

With the general definitions and notations for UDA, we can define two different measures of the gap between the source and target domains. By quantifying this domain gap, we wish to highlight in the learning bound the influence of using data from the source domain, to optimize the expected risk on the target domain.\\

The first quantity that will be used to measure the domain gap is the ideal joint error. To introduce the ideal joint error, we first define the \emph{ideal joint hypothesis} which represents the classifier that performs the best simultaneously on both domains. It is defined as:
\begin{equation}
\label{eq:idealh}
h^{*} = \argmin_{h \in \mathcal{H}} \epsilon_{S}(h) + \epsilon_{T}(h)
\end{equation}

\noindent And then, we can define the \emph{ideal joint error}.

\noindent \textbf{Definition.} If $h^{*}$ represents the ideal joint hypothesis, the \emph{ideal joint error} $\lambda$ is defined by:
\begin{equation}
\label{eq:lambda}
\lambda=\epsilon_{S}\left(h^{*}\right)+\epsilon_{T}\left(h^{*}\right)
\end{equation}
Intuitively, a large ideal joint error indicates that we cannot expect to find a hypothesis that performs well on the target and source domains. This implies that  we cannot find a classifier that performs well on the target domain only by minimizing $\epsilon_{S}$.\

Moreover, we introduce another notion to measure the domain gap in the learning bound. We refer for this to existing work on domain adaptation for classification. Particularly, we choose the definition of the $\mathcal{H} \Delta \mathcal{H}$-distance given by Ben David et al. in their paper (\cite{ben2010theory}), as it is tailored for the binary classification.\\

\noindent \textbf{Definition.} For any pair of distributions $\mathcal{P}$ and $\mathcal{Q}$ defined on $\mathcal{X}$, we define the $\mathcal{H} \Delta \mathcal{H}$-distance as:
\begin{equation}
\label{eq:hdeltah}
\begin{aligned}
d_{\mathcal{H} \Delta \mathcal{H}}( \mathcal{P}, \mathcal{Q}) = 2 \sup _{h, h^{\prime} \in \mathcal{H}}|\operatorname{Pr}_{x \sim \mathcal{P}}\left[h(x) \neq h^{\prime}(x)\right]\\-\operatorname{Pr}_{x \sim \mathcal{Q}}\left[h(x) \neq h^{\prime}(x)\right]| ,
\end{aligned}
\end{equation} where $\operatorname{Pr}_{x \sim \mathcal{P}}\left[h(x) \neq h^{\prime}(x)\right]$ (resp. $\operatorname{Pr}_{x \sim \mathcal{Q}}\left[h(x) \neq h^{\prime}(x)\right]$)  denotes the probability of ``$h(x) \neq h^{\prime}(x)$" when $x \sim \mathcal{P}$ (resp. $x \sim \mathcal{Q}$).
The $\mathcal{H} \Delta \mathcal{H}$-distance can be linked to the source and target expected risks with the following lemma:\\

\noindent \textbf{Lemma 1 (L1).} For any hypotheses $h, h^{\prime} \in \mathcal{H}$,
\begin{equation}
\label{eq:lemma1}
\left|\epsilon_{S}\left(h, h^{\prime}\right)-\epsilon_{T}\left(h, h^{\prime}\right)\right| \leq \frac{M}{2} d_{\mathcal{H} \Delta \mathcal{H}}\left(\mathcal{D}_{S}, \mathcal{D}_{T}\right),
\end{equation}
where $M > 0$ is the  ``0-M" loss bound defined in Eq.~\ref{eq:0MLoss}.

\begin{proof} Let $h, h^{\prime} \in \mathcal{H}$. We can highlight the ``0-M" loss bound by multiplying and dividing by $M > 0$ the $\mathcal{H} \Delta \mathcal{H}$ -distance expression given by definition in Eq.~\ref{eq:hdeltah}:
\begin{equation}
\Scale[0.8]{
\begin{aligned}
d_{\mathcal{H} \Delta \mathcal{H}}\left(\mathcal{D}_{S}, \mathcal{D}_{T}\right) &=2 \sup _{h, h^{\prime} \in \mathcal{H}}\left|\operatorname{Pr}_{x \sim \mathcal{D}_{S}}\left[h(x) \neq h^{\prime}(x)\right] -\operatorname{Pr}_{x \sim \mathcal{D}_{T}}\left[h(x) \neq h^{\prime}(x)\right]\right|\\
&= \frac{2}{M} \sup _{h, h^{\prime} \in \mathcal{H}}\left|M \operatorname{Pr}_{x \sim \mathcal{D}_{S}}\left[h(x) \neq h^{\prime}(x)\right]-M\operatorname{Pr}_{x \sim \mathcal{D}_{T}}\left[h(x) \neq h^{\prime}(x)\right]\right|
\end{aligned}}
\end{equation}

\noindent The expectation of the indicator function for an event is the probability of that event. Then, by rewriting the probabilities in terms of expectations, we recover the expression of the expected risks given by Eq.~\ref{eq:expected_risk}:

\begin{equation}
\Scale[0.8]{
\begin{aligned}
d_{\mathcal{H} \Delta \mathcal{H}}\left(\mathcal{D}_{S}, \mathcal{D}_{T}\right) &= \frac{2}{M} \sup _{h, h^{\prime} \in \mathcal{H}}\left| \mathop{\mathbb{E}}_{x \sim \mathcal{D}_{S}}\left[M\mathbb{1}_{h(x) \neq h^{\prime}(x)}\right]-\mathop{\mathbb{E}}_{x \sim \mathcal{D}_{T}}\left[M\mathbb{1}_{h(x) \neq h^{\prime}(x)}\right]\right| \\
&=\frac{2}{M} \sup _{h, h^{\prime} \in \mathcal{H}}\left|\epsilon_{S}\left(h, h^{\prime}\right)-\epsilon_{T}\left(h, h^{\prime}\right)\right|\\
\end{aligned}}
\end{equation}

By using the definition of the $\sup$ operator:
\begin{equation}
\begin{aligned}
d_{\mathcal{H} \Delta \mathcal{H}}\left(\mathcal{D}_{S}, \mathcal{D}_{T}\right) &\geq \frac{2}{M}\left|\epsilon_{S}\left(h, h^{\prime}\right)-\epsilon_{T}\left(h, h^{\prime}\right)\right|.
\end{aligned}
\end{equation}

\noindent Which is equivalent to the following inequality since $M>0$:

\begin{equation}
\begin{aligned}
\left|\epsilon_{S}\left(h, h^{\prime}\right)-\epsilon_{T}\left(h, h^{\prime}\right)\right| \leq \frac{M}{2} d_{\mathcal{H} \Delta \mathcal{H}}\left(\mathcal{D}_{S}, \mathcal{D}_{T}\right).
\end{aligned}
\end{equation}

\end{proof}

In this section, we defined the UDA framework, and derive from our definitions the previous Lemma 1 that will be useful afterward (to derive the Lemma 2 given by Eq.~\ref{eq:lemma2}). As this work more particularly focuses on Pseudo-Labeling UDA approach, we set a specific framework for this kind of UDA approaches in the following section.

\subsection{Modeling Pseudo-Labeling with the noisy-label framework.}
\label{sec:pseudo_label}

As motivated in Sec.~\ref{sec:state-of-the-art_udareid}, we recall that this paper focuses on the pseudo-label paradigm for UDA re-ID, i.e. learning with pseudo-labels on the target domain, since they performed the best among the UDA re-ID approaches. For this, it is necessary to choose a theoretical model in order to account for the use of pseudo-labels in the learning bounds. To this end, we propose the model of learning with noisy labels. Indeed, the Pseudo-Labeling can be seen as using a strategy in order to obtain ``artificial" labels for the unlabeled target data. Concretely, the Pseudo-Labeling process corrupts the (unseen) labels of our target samples to give pseudo-labels used as supervision during training. Therefore, the pseudo-labeled target samples can be seen as samples from a corrupted target distributions $\tilde{T} = (\mathcal{D}^T,\tilde{f}_T)$ where $\tilde{f}_T$ is a (pseudo-)labeling function  $\tilde{f}_T : \mathcal{X} \rightarrow \{-1;1\}$, which can be non-deterministic. \\
\newline
Our goal is to highlight the influence of the pseudo-labels noise on the target performance. Therefore, we follow the noise model used by Natarajan et al. in their paper (\cite{natarajan2013learning}) to derive an upper-bound on a classification expected risk: the class-conditional random noise model.  Following the class-conditional random noise model, we have: 
\begin{equation} \forall x \in \mathcal{X},
\left\{
    \begin{array}{ll}
        \rho_{-1} = \operatorname{Pr}(\tilde{f}_T(x) = 1 | f_{T}(x) = -1) \\
        \rho_{+1} = \operatorname{Pr}(\tilde{f}_T(x) = -1 | f_{T}(x) = 1)
    \end{array}
\right.
\end{equation}
with $\rho_{-1} + \rho_{+1} < 1$.
In other words, the corruption process is independent on the sample and only depends on the class. Therefore, it can be described by $\rho_{-1}$  which represents the probability that a ``-1" labeled sample 
is pseudo-labeled ``1" by $\tilde{f}_T$  
and $\rho_{+1}$ the probability that a ``1" labeled sample becomes ``-1" after Pseudo-Labeling by $\tilde{f}_T$.\\
\newline
In the presence of noise in the annotations, i.e. when using target pseudo-labeled samples, the target empirical risk $\hat{\epsilon}_{T}$ associated to $\epsilon_{T}$ becomes a biased estimate of $\epsilon_{T}$, as shown in the work of Natarajan et al. (\cite{natarajan2013learning}). That's why, following their Lemma 1 (\cite{natarajan2013learning}), we define the \emph{corrected loss function} $\tilde{\mathcal{L}}$:
\begin{equation}
\label{eq:corrected_loss}
\forall y,y' \in \{-1;1\},
\tilde{\mathcal{L}}(y, y') = \frac{(1-\rho_{-y'})\mathcal{L}(y, y')-\rho_{y'}\mathcal{L}(y,-y')}{1-\rho_{+1} -\rho_{-1}}.
\end{equation} where $\rho_{y'} = \rho_{-1}$ if $y' = -1$ and $\rho_{y'} = \rho_{+1}$ if $y' = 1$.
Therefore, also according to their Lemma 1 and following the previous notation: 
\begin{equation}
\label{eq:corrected_risk}
\forall h \in \mathcal{H}, \mathbb{E}_{x \sim \mathcal{D}^T}[ \tilde{\mathcal{L}}(h(x),\tilde{f}_T(x))] = \epsilon_{T}(h).
\end{equation}
In the rest of the paper, we denote by $\tilde{\epsilon}_{T}$ the \emph{target empirical corrected risk}, associated to the corrected loss function defined in Eq.~\ref{eq:corrected_loss}, and computed with target pseudo-labeled samples. It is an unbiased estimate of $\epsilon_{T}$ according to Eq.~\ref{eq:corrected_risk}.




\section{Establishing a new Learning Bound for Pseudo-Labeling UDA re-ID.}
\label{sec:bound}

To integrate the source in the training stage, we assume that we optimize a convex weighting of the source empirical risk associated to $\epsilon_{S}$ and corrected target empirical risks, that we call the \emph{Source-guided empirical risk with target pseudo-labels} $\hat{\epsilon}_{\alpha}$. For this purpose, we have $m$ training samples in total, of which $\beta m$ are i.i.d pseudo-labeled target samples $(x^{T}_i,\tilde{y}^{T}_i)_{1 \leq i \leq \beta m}$ from $\tilde{T}$ and $(1-\beta) m$ are labeled source samples$(x^{S}_i,y^{S}_i)_{1 \leq i \leq (1-\beta) m}$ from $S$, $\beta \in [0,1]$. With these samples, $\hat{\epsilon}_{S}$, $\tilde{\epsilon}_{T}$ can be expressed as, for any $h \in \mathcal{H}$:

\begin{equation}
\label{eq:empiricalrisks}
\left\{
    \begin{array}{ll}
        \tilde{\epsilon}_{T}(h) &= \frac{1}{\beta m}\sum_{i=1}^{\beta m} \tilde{\mathcal{L}}(h(x^{T}_i),\tilde{y}^{T}_i)\\ &= \frac{1}{\beta m}\sum_{i=1}^{\beta m} \tilde{\mathcal{L}}(h(x^{T}_i),\tilde{f}_T(x^{T}_i)) \\
        \hat{\epsilon}_{S}(h) &= \frac{1}{(1-\beta)m}\sum_{i=1}^{(1-\beta) m}\mathcal{L}(h(x^{S}_i),y^{S}_i)\\& = \frac{1}{(1-\beta)m}\sum_{i=1}^{(1-\beta) m}\mathcal{L}(h(x^{S}_i),f_S(x^{S}_i))
    \end{array}
\right.
\end{equation}

And then $\hat{\epsilon}_{\alpha}$ can be expressed as:
\begin{equation}
\label{eq:empiricalsourceguidedrisk}
\hat{\epsilon}_{\alpha} = \alpha \tilde{\epsilon}_{T} + (1-\alpha)\hat{\epsilon}_{S}, \alpha \in [0,1]
\end{equation}.
We also define the quantity $\epsilon_{\alpha}$ given by:
\begin{equation}
\label{eq:sourceguidedrisk}
\epsilon_{\alpha} = \alpha \epsilon_{T} + (1-\alpha)\epsilon_{S}, \alpha \in [0,1].
\end{equation}


Therefore, $\hat{h} = \arg\min_{h \in\mathcal{H}} \hat{\epsilon}_{\alpha}(h)$ will be our model learned by Source-guided Pseudo-Labeling, that is to say by minimizing the Source-guided empirical risk with target pseudo-labels $\hat{\epsilon}_{\alpha}$. In the UDA setting, we want this model to have the best re-ID performance on the target domain, that is to say we want to reduce  $\epsilon_{T}(\hat{h})$. Therefore, we would like to establish an upper-bound on $\epsilon_{T}(\hat{h})$. As Ben David et al. in their paper (\cite{ben2010theory}), to establish a learning bound on $\epsilon_{T}(\hat{h})$, we proceed in three steps:
\begin{itemize}
\item Linking $\epsilon_{\alpha}$ to the target expected risk $\epsilon_{T}$, with the following Lemma 2.
\item Linking $\hat{\epsilon}_{\alpha}$ to $\epsilon_{\alpha}$ with the following Lemma 3.
\item Using Lemma 2 and Lemma 3 to build the desired upper-bound on $\epsilon_{T}(\hat{h})$
\end{itemize}

\subsection{Preliminary lemmas to establish the Source-guided Pseudo-Labeling upper-bound for UDA re-ID.}

We first link $\epsilon_{\alpha}$ to the target expected risk $\epsilon_{T}$ (that we wish to minimize) with the following lemma. 

\noindent \textbf{Lemma 2 (L2).} Let $h$ be a classifier in $\mathcal{H}$. Then
\begin{equation}
\label{eq:lemma2}
\left|\epsilon_{\alpha}(h)-\epsilon_{T}(h)\right| \leq(1-\alpha)\left(\frac{M}{2} d_{\mathcal{H} \Delta \mathcal{H}}\left(\mathcal{D}_{S}, \mathcal{D}_{T}\right)+\lambda\right) .
\end{equation}

\begin{proof} Let $h \in \mathcal{H}$ and $h^*$ be the ideal joint hypothesis defined in Eq.~\ref{eq:idealh}. By definition of $\epsilon_{\alpha}$ (Eq.~\ref{eq:sourceguidedrisk}), we can write:

\begin{equation}
\begin{aligned}
\left|\epsilon_{\alpha}(h)-\epsilon_{T}(h)\right| &= \left|\alpha \epsilon_{T}(h) + (1-\alpha)\epsilon_{S}(h) -\epsilon_{T}(h)\right|\\
&= (1-\alpha)\left|\epsilon_{S}(h)-\epsilon_{T}(h)\right|\\
&= (1-\alpha)|\epsilon_{S}(h)-\epsilon_{S}(h, h^{*})+\epsilon_{S}(h, h^{*})-\epsilon_{T}(h, h^{*})\\&+\epsilon_{T}(h, h^{*})-\epsilon_{T}(h)|\\
\end{aligned}
\end{equation}

\noindent Applying the triangular inequality property of $\left| . \right|$, we obtain:

\begin{equation}
\begin{aligned}
\left|\epsilon_{\alpha}(h)-\epsilon_{T}(h)\right| &\leq (1-\alpha)[\left|\epsilon_{S}(h)-\epsilon_{S}\left(h, h^{*}\right)\right|+|\epsilon_{S}\left(h, h^{*}\right)\\&-\epsilon_{T}\left(h, h^{*}\right)|+\left|\epsilon_{T}\left(h, h^{*}\right)-\epsilon_{T}(h)\right|]
\end{aligned}
\end{equation}

\noindent Using the triangular inequality property of $\epsilon_{S}(.,.)$ and $\epsilon_{T}(.,.)$ , the inequality becomes:
\begin{equation}
\begin{aligned}
\left|\epsilon_{\alpha}(h)-\epsilon_{T}(h)\right| &\leq(1-\alpha)[\epsilon_{S}\left(h^{*}\right)+\left|\epsilon_{S}\left(h, h^{*}\right)-\epsilon_{T}\left(h, h^{*}\right)\right|\\&+\epsilon_{T}\left(h^{*}\right)]
\end{aligned}
\end{equation}

\noindent And then by applying Lemma 1 (Eq.~\ref{eq:lemma1}) and by definition of $\lambda$ (Eq.~\ref{eq:lambda}):
\begin{equation}
\begin{aligned}
\left|\epsilon_{\alpha}(h)-\epsilon_{T}(h)\right|
&\leq(1-\alpha)\left(\frac{M}{2} d_{\mathcal{H} \Delta \mathcal{H}}\left(\mathcal{D}_{S}, \mathcal{D}_{T}\right)+\lambda\right).
\end{aligned}
\end{equation}
\end{proof}

We can also link the empirical risk $\hat{\epsilon}_{\alpha}$ (Eq.~\ref{eq:empiricalsourceguidedrisk}), to the expected risk $\epsilon_{\alpha}$ (Eq.~\ref{eq:sourceguidedrisk}), with the following lemma. 

\noindent \textbf{Lemma 3 (L3).} For any $\mu>0$:
\begin{equation}
\operatorname{Pr}\left[\left|\hat{\epsilon}_{\alpha}(h)-\epsilon_{\alpha}(h)\right| \geq \mu\right] \leq 2 \exp \left(\frac{-2 m \mu^{2}}{\frac{4M^2 \alpha^{2}}{\beta(1-\rho_{+1} -\rho_{-1})^2}+\frac{M^2(1-\alpha)^{2}}{1-\beta}}\right),
\end{equation}

As Ben David et al. for their theorem 3, we refer to Anthony and
Bartlett (\cite{anthony1999neural}) for the detailed classical steps to derive the following VC-dimension upper-bound, using the inequality concentration from our Lemma 3. If $d$ is the VC-dimension of $\mathcal{H}$,  with a probability $\delta$ ($0 \leq \delta \leq 1$) on the drawing of the samples, we have:

\begin{equation}
\label{eq:lemma3}
\Scale[0.8]{
\left|\epsilon_{\alpha}(h)-\hat{\epsilon}_{\alpha}(h)\right| \leq 2 \sqrt{\frac{4M^2 \alpha^2}{\beta(1-\rho_{+1} -\rho_{-1})^2}+\frac{M^{2}(1-\alpha)^{2}}{1-\beta}} \sqrt{\frac{2 d \log (2(m+1))+2 \log \left(\frac{8}{\delta}\right)}{m}}.
}
\end{equation}

\noindent Before giving the Lemma 3's proof, we recall the Hoeffding's inequality:\\
\noindent \textbf{Hoeffding's inequality}. If $X_{1}, \ldots, X_{n}$ are independent random variables with $a_{i} \leq X_{i} \leq b_{i}$ for all $i$, then for any $\mu>0$,
\begin{equation}
\label{eq:hoeffding}
\operatorname{Pr}[|\bar{X}-E[\bar{X}]| \geq \mu] \leq 2 \exp\left(\frac{-2 n^{2} \mu^{2}}{ \sum_{i=1}^{n}\left(\operatorname{range(X_{i})}\right)^{2}}\right),
\end{equation}
where $\bar{X}=\left(X_{1}+\cdots+X_{n}\right) / n$ and $\forall 1 \leq i \leq n, \operatorname{range(X_{i})} = b_{i}-a_{i}.$

\begin{proof} Let $h \in \mathcal{H}$ and $X_{1}, \ldots, X_{\beta m}$  be some random variables taking the values:\begin{equation}
\frac{\alpha}{\beta}\tilde{\mathcal{L}}(h(x_{1}),\tilde{f}_T(x_{1})), \ldots, \frac{\alpha}{\beta}\tilde{\mathcal{L}}(h(x_{\beta m}),\tilde{f}_T(x_{\beta m}))\end{equation}
for the random variables $x_1 \ldots x_{\beta m}$ associated to the generation of the $\beta m$ samples from the pseudo-labeled target domain $\tilde{T}$. Similarly, let $X_{\beta m+1}, \ldots, X_{m}$ be some random variables taking the value:
\begin{equation}
\frac{1-\alpha}{1-\beta}\mathcal{L}(h(x_{\beta m + 1}),f_{S}(x_{\beta m + 1})), \ldots, \frac{1-\alpha}{1-\beta}\mathcal{L}(h(x_{m}),f_{S}(x_{m}))\end{equation}
for the random variables  $x_{\beta m + 1}, \ldots, x_{m}$ associated to the generation of the $(1-\beta) m$ samples from the source domain $S$. 

By definition of $\hat{\epsilon}_{\alpha}$ (Eq.~\ref{eq:empiricalsourceguidedrisk}), we can write:
\begin{equation}\hat{\epsilon}_{\alpha}(h) = \alpha \tilde{\epsilon}_{T}(h)+(1-\alpha) \hat{\epsilon}_{S}(h)\end{equation}

And by definition of $\tilde{\epsilon}_{T}$ and $\hat{\epsilon}_{S}$ (Eq.~\ref{eq:empiricalrisks}), we have:

\begin{equation}
\Scale[0.8]{
\begin{aligned}
\hat{\epsilon}_{\alpha}(h) &= \alpha \frac{1}{\beta m}\sum_{i=1}^{\beta m}\tilde{\mathcal{L}}(h(x_{i}),\tilde{f}_T(x_{i})) + (1-\alpha)\frac{1}{(1-\beta)m}\sum_{i=\beta m+1}^{ m}\mathcal{L}(h(x_{i}),f_{S}(x_{i}))\\
&= \frac{1}{m} \left( \sum_{i=1}^{\beta m}\frac{\alpha}{\beta}\tilde{\mathcal{L}}(h(x_{i}),\tilde{f}_T(x_{i})) + \sum_{i=\beta m+1}^{ m}\frac{(1-\alpha)}{(1-\beta)}\mathcal{L}(h(x_{i}),f_{S}(x_{i})) \right)
\end{aligned}
}
\end{equation}

Then we can write, by definition of $X_1, \ldots, X_{m}$:
\begin{equation}
\hat{\epsilon}_{\alpha}(h) = \frac{1}{m} \sum_{i=1}^{m} X_{i}\end{equation}

Using the linearity of expectations, we have:
\begin{equation}
\label{eq:expected_alpha}
\begin{aligned}
\mathbb{E}_{x}[\hat{\epsilon}_{\alpha}(h)] &= \frac{1}{m} (\frac{\alpha}{\beta}\sum_{i=1}^{\beta m}\mathbb{E}_{x_i \sim  \mathcal{D}_T}[\tilde{\mathcal{L}}(h(x_{i}),\tilde{f}_T(x_{i}))] \\&+ \frac{(1-\alpha)}{(1-\beta)}\sum_{i=\beta m+1}^{m}\mathbb{E}_{x_i \sim  \mathcal{D}_S}[\mathcal{L}(h(x_{i}),f_{S}(x_{i}))])
\end{aligned}
\end{equation}

According to Eq.~\ref{eq:expected_risk}: 
\begin{equation}
\begin{aligned}
\mathbb{E}_{x}[\hat{\epsilon}_{\alpha}(h)] &= \frac{1}{m}\left(\frac{\alpha}{\beta}\beta m  \epsilon_{T}(h)+ \frac{1-\alpha}{1-\beta}(1-\beta)m \epsilon_{S}(h)\right) \\
&=\alpha \epsilon_{T}(h)+(1-\alpha) \epsilon_{S}(h)
\end{aligned}
\end{equation}

And then by definition of $\epsilon_{\alpha}$ (Eq.~\ref{eq:sourceguidedrisk}):

\begin{equation}
\begin{aligned}
\mathbb{E}_{x}[\hat{\epsilon}_{\alpha}(h)] = \epsilon_{\alpha}(h)
\end{aligned}
\end{equation}

Moreover, by definition of $\tilde{\mathcal{L}}$ (Eq.~\ref{eq:corrected_loss}) and $\mathcal{L}$ (Eq.~\ref{eq:0MLoss}), we have:
\begin{equation}
\forall y,y' \in \{-1;1\}, 
\left\{
    \begin{array}{ll}
       \frac{-M}{1-\rho_{+1}-\rho_{-1}} \leq \tilde{\mathcal{L}}(y,y') \leq \frac{M}{1-\rho_{+1}-\rho_{-1}} \\
        0 \leq \mathcal{L}(y,y') \leq M
    \end{array}
\right.
\end{equation}

Therefore, we can say that $X_{1}, \ldots, X_{\beta m} \in[-\frac{M \alpha}{\beta(1-\rho_{+1} -\rho_{-1})}, \frac{M \alpha}{\beta(1-\rho_{+1} -\rho_{-1})} ]$ and $X_{\beta m+1}, \ldots, X_{m} \in
[0,\frac{M(1-\alpha)}{(1-\beta)}]$. And then, we have:
\begin{equation}
\label{eq:range}
\operatorname{range}(X_{i}) = 
\left\{
    \begin{array}{ll}
    \frac{2M \alpha}{\beta(1-\rho_{+1} -\rho_{-1})}, 1 \leq i \leq \beta m \\
        \frac{M(1-\alpha)}{1-\beta}, \beta m + 1 \leq i \leq m
    \end{array}
\right.
\end{equation}

According to Eq.~\ref{eq:expected_alpha},  $\forall \mu > 0$:

\begin{equation}
\begin{aligned}
\operatorname{Pr}\left[\left|\hat{\epsilon}_{\alpha}(h)-\epsilon_{\alpha}(h)\right| \geq \mu\right] &= \operatorname{Pr}\left[\left|\hat{\epsilon}_{\alpha}(h)-\mathbb{E}_{x}[\hat{\epsilon}_{\alpha}(h)]\right| \geq \mu\right]
\end{aligned}
\end{equation}
Then by applying the Hoeffding's inequality (Eq.~\ref{eq:hoeffding}) to $\hat{\epsilon}_{\alpha}(h)$:
\begin{equation}
\Scale[0.8]{
\begin{aligned}
\operatorname{Pr}\left[\left|\hat{\epsilon}_{\alpha}(h)-\epsilon_{\alpha}(h)\right| \geq \mu\right] & \leq 2 \exp \left(\frac{-2 m^{2} \mu^{2}}{\sum_{i=1}^{m} \operatorname{range}\left(X_{i}\right)^{2}}\right) \\
& \leq 2 \exp \left(\frac{-2 m^{2} \mu^{2}}{\sum_{i=1}^{\beta m} \operatorname{range}\left(X_{i}\right)^{2}+ \sum_{i=\beta m + 1}^{m} \operatorname{range}\left(X_{i}\right)^{2}}\right)
\end{aligned}
}
\end{equation}
Then according to Eq.~\ref{eq:range}:
\begin{equation}
\Scale[0.8]{
\begin{aligned}
\operatorname{Pr}\left[\left|\hat{\epsilon}_{\alpha}(h)-\epsilon_{\alpha}(h)\right| \geq \mu\right] &\leq 2 \exp \left(\frac{-2 m^{2} \mu^{2}}{\beta m\left(\frac{2M \alpha}{\beta(1-\rho_{+1} -\rho_{-1})}\right)^{2}+(1-\beta) m\left(\frac{M(1-\alpha)}{1-\beta}\right)^{2}}\right) \\
&\leq 2 \exp \left(\frac{-2 m \mu^{2}}{\frac{4M^2 \alpha^{2}}{\beta(1-\rho_{+1} -\rho_{-1})^2}+\frac{M^{2}(1-\alpha)^{2}}{1-\beta}}\right).
\end{aligned}
}
\end{equation}
\end{proof}

\subsection{A new learning bound for Pseudo-Labeling UDA.}

Using the previous notation, we define the \emph{ideal target hypothesis} $h^{*}_{T} = \arg\min_{h \in\mathcal{H}} \epsilon_{T}(h)$. The upper-bound on $\epsilon_{T}(\hat{h})$ can be established. \\

\noindent \textbf{Theorem.} With a probability $1-\delta$ ($0 \leq \delta \leq 1$) on the drawing of the samples, we have:\\
\begin{equation}
\Scale[0.75]{
\begin{aligned}
\label{eq:bound}
\epsilon_{T}(\hat{h}) {}  & \leq \epsilon_{T}(h^{*}_{T}) + 4M \overbrace{ \sqrt{\frac{4\alpha^2}{\beta (1-\rho_{+1}-\rho_{-1})^2} + \frac{1-\alpha^2}{1-\beta}}}^{\mathcal{N}} \overbrace{\sqrt{\frac{2d\log(2(m+1)) +2\log(\frac{8}{\delta})}{m}}}^{\mathcal{C}} \\
 & + 
2\underbrace{(1-\alpha)(\frac{M}{2}{d}_{\mathcal{H} \Delta \mathcal{H}}(\mathcal{D}_{S}, \mathcal{D}_{T}) + \lambda}_{\mathcal{D}\mathcal{D}}).
\end{aligned}
}
\end{equation}

$\mathcal{N}$, $\mathcal{C}$ and $\mathcal{D} \mathcal{D}$ correspond to noteworthy terms that will be discussed in the next section (Sec.~\ref{sec:theory_to_practice}) to get insight from the upper-bound.

\begin{proof} 
Following the previous notations, let $0 \leq \delta \leq 1$.
We first use the Lemma 2 (Eq.~\ref{eq:lemma2}) to bound $\epsilon_{T}(\hat{h})$:
\begin{equation}
\epsilon_{T}(\hat{h}) \leq \epsilon_{\alpha}(\hat{h})+(1-\alpha)\left(\frac{M}{2} d_{\mathcal{H} \Delta \mathcal{H}}\left(\mathcal{D}_{S}, \mathcal{D}_{T}\right)+\lambda\right)
\end{equation}
Then by using the upper-bound on $\epsilon_{\alpha}(\hat{h})$ derived from Lemma 3 (Eq.~\ref{eq:lemma3}), we have
\begin{equation}
\Scale[0.75]{
\begin{aligned}
\epsilon_{T}(\hat{h}) \leq & \hat{\epsilon}_{\alpha}(\hat{h})+2 \sqrt{\frac{4M^2 \alpha^2}{\beta(1-\rho_{+1} -\rho_{-1})^2}+\frac{M^{2}(1-\alpha)^{2}}{1-\beta}} \sqrt{\frac{2 d \log (2(m+1))+2 \log \left(\frac{8}{\delta}\right)}{m}} \\
&+(1-\alpha)\left(\frac{M}{2} d_{\mathcal{H} \Delta \mathcal{H}}\left(\mathcal{D}_{S}, \mathcal{D}_{T}\right)+\lambda \right) \\
\end{aligned}
}
\end{equation}
Since $\hat{h}=\arg \min _{h \in \mathcal{H}} \hat{\epsilon}_{\alpha}(h)$, we have $\hat{\epsilon}_{\alpha}(\hat{h}) \leq \hat{\epsilon}_{\alpha}(h_{T}^{*})$, and therefore:
\begin{equation}
\Scale[0.75]{
\begin{aligned}
\epsilon_{T}(\hat{h}) \leq & \hat{\epsilon}_{\alpha}\left(h_{T}^{*}\right)+2M \sqrt{\frac{4\alpha^{2}}{\beta(1-\rho_{+1} -\rho_{-1})^2}+\frac{(1-\alpha)^{2}}{1-\beta}} \sqrt{\frac{2 d \log (2(m+1))+2 \log \left(\frac{8}{\delta}\right)}{m}} \\
&+(1-\alpha)\left(\frac{M}{2} d_{\mathcal{H} \Delta \mathcal{H}}\left(\mathcal{D}_{S}, \mathcal{D}_{T}\right)+\lambda\right)
\end{aligned}
}
\end{equation}
And then, by using the Lemma 3 (Eq.~\ref{eq:lemma3}) to bound $\hat{\epsilon}_{\alpha}(h_{T}^{*})$, we have:
\begin{equation}
\Scale[0.75]{
\begin{aligned}
\epsilon_{T}(\hat{h}) \leq &  \epsilon_{\alpha}\left(h_{T}^{*}\right)+4M \sqrt{\frac{4\alpha^{2}}{\beta(1-\rho_{+1} -\rho_{-1})^2}+\frac{(1-\alpha)^{2}}{1-\beta}} \sqrt{\frac{2 d \log (2(m+1))+2 \log \left(\frac{8}{\delta}\right)}{m}} \\
&+(1-\alpha)\left(\frac{M}{2} d_{\mathcal{H} \Delta \mathcal{H}}\left(\mathcal{D}_{S}, \mathcal{D}_{T}\right)+\lambda\right)
\end{aligned}
}
\end{equation}
Finally, by using the Lemma 2 (Eq.~\ref{eq:lemma2}) to upper-bound $\epsilon_{\alpha}(h_{T}^{*})$, we have:
\begin{equation}
\Scale[0.75]{
\begin{aligned}
\epsilon_{T}(\hat{h}) \leq & \epsilon_{T}\left(h_{T}^{*}\right)+4M \sqrt{\frac{4\alpha^{2}}{\beta(1-\rho_{+1} -\rho_{-1})^2}+\frac{(1-\alpha)^{2}}{1-\beta}} \sqrt{\frac{2 d \log (2(m+1))+2 \log \left(\frac{8}{\delta}\right)}{m}} \\
&+2(1-\alpha)\left(\frac{M}{2} d_{\mathcal{H} \Delta \mathcal{H}}\left(\mathcal{D}_{S}, \mathcal{D}_{T}\right)+\lambda\right)
\end{aligned}
}
\end{equation}
\end{proof}

To the best of our knowledge, this bound is entirely new, and takes into account the interactions between the source data and the pseudo-annotated target data. Even if the overall upper-bound establishment is inspired by Ben David et al. (\cite{ben2010theory}), we recall that theirs does not take into account the use of pseudo-labels for the target data nor the loss bound given by $M$. Similarly, Natarajan et al. (\cite{natarajan2013learning}) does not take into account the use of source data for learning. In the following Sec.~\ref{sec:theory_to_practice}, we propose an interpretation of this bound in order to get more insight on Source-guided Pseudo-Labeling learning. Then, we derive good practices from it.

\section{Interpretation of the bound and derived good practices.}
\label{sec:theory_to_practice}


Taking into account errors in the pseudo-labels, as well as the domain discrepancy between source and target, leads to a new learning upper-bound on Source-guided Pseudo-Labeling for UDA re-ID given by Eq.~\ref{eq:bound} in Sec.~\ref{sec:bound}. This bound could be used to find the best hyperparameter $\alpha$ to weight the source-guidance in the Source-guided Pseudo-Labeling empirical risk $\hat{\epsilon_{\alpha}}$ (Eq.~\ref{eq:empiricalsourceguidedrisk}) i.e. to find an optimal solution $\alpha^*$ minimizing the upper-bound as a function of $\alpha$. This optimization has been solved by Ben David et al. (\cite{ben2010theory}) with their upper-bound for binary classification (with target ground-truth labels), to find the optimal mixing value between source and target in $\hat{\epsilon}_{\alpha}$. However, in the case of our bound, $\alpha^*$ would eventually depend on the noise probabilities $\rho_{-1}$ and $\rho_{+1}$. These values are not known in practice. Therefore, the interest of looking for such a solution $\alpha$ is limited from a practical perspective. However, it is still possible to get insight about Source-guided Pseudo-Labeling for UDA re-ID by analyzing the different terms of the bound: $\mathcal{N}$, $\mathcal{C}$ and $\mathcal{D}\mathcal{D}$ resp. for the Noise term, the Complexity term, and the Domain Discrepancy term.
Therefore, we consider $\alpha$ as a hyperparameter specified before training the model. Then, we propose to analyze this bound by looking at the influence on the target performance of its key elements: the complexity term, the noise term and the domain discrepancy term. This analysis contributes to answering the question of how to better use the source data during training with pseudo-labels, and to deduce general good practices to follow.

\subsection{Noise term: $\mathcal{N}$.}

The noise term $\mathcal{N}$ of our new learning bound for Pseudo-Labeling UDA re-ID (Eq.~\ref{eq:bound} in Sec.~\ref{sec:bound}) involves the noise probabilities $\rho_{-1}$ and $\rho_{+1}$, as well as $\alpha$ which describes the weight put on the (noisy) target data in the Source-guided Pseudo-Labeling empirical risk $\hat{\epsilon}_{\alpha}$ (Eq.~\ref{eq:empiricalsourceguidedrisk}) minimized during training. Intuitively, this term represents the impact of errors in the pseudo-labels on the re-ID performance: the higher $\rho_{-1}$ and $\rho_{+1}$, the higher $\mathcal{N}$. However, increasing $\mathcal{N}$ increases the upper-bound, and therefore is more likely to degrade the target re-ID performance $\epsilon_{T}(\hat{h})$. \\
While $\rho_{-1}$ and $\rho_{+1}$ are unknown, it is possible to reduce them in order to reduce $\mathcal{N}$. Indeed, these probabilities could be estimated by the proportions of mislabeled pairs of pseudo-labeled data. Even if we cannot directly compute these quantities without the ground-truth target labels, it is possible to reduce them.  For example, they can be reduced with a pseudo-label refinement strategy or with a strategy of filtering out the outliers (mis-pseudo-labeled data) in the target training set.

\subsection{Complexity term: $\mathcal{C}$.}

The complexity term $\mathcal{C}$ of our new learning bound for Pseudo-Labeling UDA re-ID (Eq.~\ref{eq:bound} in Sec.~\ref{sec:bound}) involves the VC-dimension $d$ of $\mathcal{H}$ measuring the complexity of our selected set of models. Moreover, it also involves the total number $m$  of training data, which includes the source and target samples. Intuitively, the complexity term measures how well the class of hypothesis can memorize the training dataset given its number of samples.
According to the bound, reducing it should also reduce the expected target risk $\epsilon_{T}(\hat{h})$ of our learned model $\hat{h}$.
To reduce it, two options are available. On the one hand, we could reduce the VC-dimension $d$ of the hypothesis class: this is what is classically done in machine learning using regularization on the learned model parameters such as weight decay.\\ The other option is to increase $m$, and for that we have to use as much data as possible during the training. $m$ can be increased artificially by data-augmentation, but also by including all the source data for the training with the pseudo-labeled target data.\\
\noindent \textbf{Interactions with $\mathcal{N}$.}
$\mathcal{C}$ interacts multiplicatively with $\mathcal{N}$. This means that reducing $\mathcal{C}$ limits the negative impact of a high $\mathcal{N}$ by multiplication. More concretely, this indicates that Model Regularization can help to reduce the negative impact of noise on the final UDA re-ID performance. Particularly, using the source data, for Pseudo-Labeling, would allow to reduce the impact of the noise in the pseudo-labels, by reducing overfitting of the training data.

\subsection{Domain Discrepancy term: $\mathcal{D} \mathcal{D}$.}

The Domain Discrepancy term $\mathcal{D} \mathcal{D}$ of our new learning bound for Pseudo-Labeling UDA re-ID (Eq.~\ref{eq:bound} in Sec.~\ref{sec:bound}) can be decomposed into the the $\mathcal{H} \Delta \mathcal{H}$-distance between the source and target domains $d_{\mathcal{H} \Delta \mathcal{H}}\left(\mathcal{D}_{S}, \mathcal{D}_{T}\right)$, the ideal joint error $\lambda$, and the quantity $1-\alpha$ representing the weight put on the source samples in the empirical risk $\hat{\epsilon}_{\alpha}$ (Eq.~\ref{eq:empiricalsourceguidedrisk}). 
As we recall, the best $\alpha$ cannot be easily estimated in a UDA problem. It is also impossible to act on the ideal joint error $\lambda$ that is set given the class of hypothesis $\mathcal{H}$ and the source and target domains.  Therefore, to reduce the expected target risk $\epsilon_{T}(\hat{h})$, $d_{\mathcal{H} \Delta \mathcal{H}}\left(\mathcal{D}_{S}, \mathcal{D}_{T}\right)$ should remain as low as possible. For this, in practice, the Domain Discrepancy between the source and target domains is penalized to learn the UDA re-ID model i.e. Domain Alignment is performed.

\noindent \textbf{Interactions with $\mathcal{N}$.}
$\mathcal{D} \mathcal{D}$ interacts with $\mathcal{N}$, indirectly through a trade-off controlled by the $\alpha$ term. More specifically, the more the training relies on the source-guidance ($\alpha$ ``close to" 0, which reduces the negative impact of $\rho_{-1}$ and $\rho_{+1}$ in $\mathcal{N}$), the more $\mathcal{D} \mathcal{D}$ increases. Therefore, the more the training relies on source-guidance, the more important it is to reduce the Domain Discrepancy by Domain Alignment in order not to degrade the UDA re-ID performance.

\subsection{Loss Bound: $M$.}

The loss bound $M$ of our new learning bound for Pseudo-Labeling UDA re-ID (Eq.~\ref{eq:bound} in Sec.~\ref{sec:bound}) is highlighted by the use of the ``0-M" loss. It can increase multiplicatively the negative impact of $\mathcal{N}$ and $d_{\mathcal{H} \Delta \mathcal{H}}\left(\mathcal{D}_{S}, \mathcal{D}_{T}\right)$ in $\mathcal{D} \mathcal{D}$. $M$ should therefore be controlled and limited, for example by favouring the use of Bounded Loss to learn re-ID.

\begin{table*}[t]
\centering
\caption{Summary of the relationships between the theoretical analysis conducted throughout the paper and the implementation of good practices in Pseudo-Labeling UDA re-ID frameworks. Examples refer to existing state-of-the-art solutions to enforce good practices in the framework. \label{table:summarygp}}
\resizebox{\linewidth}{!}{
\begin{tabular}{@{}ccc@{}}
\toprule
Theory  & Good practices   & Examples of implementations   \\ 
\midrule
Noise Reduction ( $ \searrow \mathcal{N}$ )            & Outlier Filtering (Pseudo-Label Refinery)   & DBSCAN, Asymetric Co-Teaching, Online Outlier Filtering, etc   \\
\hdashline
Overfitting Reduction  ( $ \searrow \mathcal{C}$ )       & \makecell[c]{ Model Regularization\\ Source-guided Learning} & Mean Teaching, Feature Memory Bank, etc\\ 
\hdashline
Domain Discrepancy Reduction ( $ \searrow \mathcal{D}\mathcal{D}$ )  &  Domain Alignment  & Maximum Mean Discrepancy, Adversarial Domain Adaptation, etc\\ 
\hdashline
Bound Limitation  ( $ M < +\infty$  )        &  Bounded re-ID Loss & Loss Thresholding/Normalization, Triplet Loss with Normalized Features, etc \\
\bottomrule
\end{tabular}
}
\end{table*}

\subsection{The deduced good practices.}

Our bound analysis allows us to understand more clearly the Pseudo-Labeling training paradigm using the source samples. Furthermore, it also allows to deduce general good practices to improve cross-domain performance of UDA re-ID, detailed hereafter:
\begin{itemize}
\item \textbf{Source-guided Learning with Domain Alignment}: it consists in reducing overfitting by using the labeled source data for re-ID feature learning, particularly of the pseudo-label errors. It should be performed jointly with Domain Alignment, which constrains the feature encoder to align the source and target domains in the feature space, in order to alievate the Domain Discrepancy negative impact on the performance.
\item \textbf{Bounded Loss}: it consists in reducing the amplification of the Domain Discrepancy and Complexity terms by $M$, by using a Bounded Loss for re-ID feature learning.
\item \textbf{Outlier Filtering}: it consists in reducing the impact of the noise term by filtering outliers in pseudo-labeled target samples.
\item \textbf{Model Regularization}: it consists in reducing noise overfitting by regularizing the model.
\end{itemize}
According to our theoretical analysis, following these good practices should improve performance on the target domain and make the best use of source data when training with pseudo-labels. In Fig.~\ref{fig:good_practice}, we illustrate the classical Pseudo-Labeling cyclic training, as well was, a Pseudo-Labeling training where all good practices are followed: Source-guided Pseudo-Label training with good practices. Good practices are general and represented in green on the figure. In practice, these good practices can be implemented in various ways that we discuss in the next section.

\section{Experimental setting}
\label{sec:experiments}

This section aims at validating by experiments good practices derived from the theoretical learning bound in Sec.~\ref{sec:theory_to_practice}. For this, we propose different ways to implement good practices in a Pseudo-Labeling method of interest in Sec.~\ref{sec:exp_gp}. Then, we introduce a set of various state-of-the-art baselines in Sec.~\ref{sec:baselines}, that follow different good practices, to which we will implement the missing ones. In Sec~\ref{sec:exp_details}, details for experiments are given. Finally, Sec~\ref{sec:datasets} introduces the cross-dataset benchmarks to evaluate experiments on good practices.

\noindent \subsection{Implementing good practices into a Pseudo-Labeling framework.}
\label{sec:exp_gp}

In order to experimentally validate good practices derived from  the theory for UDA re-ID Pseudo-Labeling, they have to be implemented into Pseudo-Labeling methods of interest. Hereafter, different possible ways to implement them in Pseudo-Labeling UDA re-ID methods are proposed. The links between the theory, good practices and their implementations are summarized in Tab.~\ref{table:summarygp}.\\

\noindent \textbf{Source-guided Learning with Domain Alignment.} 
Source-guided Learning consists in learning the re-ID model with the source samples, in addition to the target samples. Simultaneously with learning with the source data, good practices include Domain Alignment. More specifically, to establish our bound, we modeled the re-ID problem as a verification task on pairs of images. Therefore, the binary classifier takes as input a vector measuring the similarity between the pair of image features. We call this space the similarity space. Domain Alignment should be performed in this space according to the theory.\\
for Learning, the hyperparameter $\alpha$ must be set in the empirical loss function $\hat{\epsilon_{\alpha}}$ (Eq.~\ref{eq:empiricalsourceguidedrisk}). Without a priori knowledge, we arbitrarily put the same weight on the source and target samples contribution in the Source-guided Pseudo-Labeling re-ID loss function, i.e. $\alpha = 0.5$ for all the experiments. Indeed, setting $\alpha$ to $0.5$ is generally done in UDA re-ID frameworks using the source data (\cite{ge2020self, dub2020, dubourvieux2021improving}). For Domain Alignment in the similarity space, minimization of the Maximum Mean Discrepancy (MMD) (\cite{gretton2012kernel} has already been used for UDA re-ID in other works (\cite{mekhazni2020unsupervised,dubourvieux2021improving}). Note that Domain Alignment can be implemented differently, like for instance with a 2-layer Domain Adversarial Neural Network (DANN (\cite{ganin2016domain})). The Optimal Transport can also be considered to perform Domain Alignment based on the Sinkhorn algorithm (\cite{cuturi2013sinkhorn}) (with a regularization parameter set to 1). Following some existing works (\cite{mekhazni2020unsupervised,dubourvieux2021improving}) for UDA re-ID, MMD is chosen with the same kernel settings in the rest of the paper. 

\noindent \textbf{Outlier Filtering.} Outlier Filtering aims at improving the number of correctly pseudo-labeled samples, by pseudo-label refinery or by discarding erroneous pseudo-labeled samples from the training stages. It can be done in an offline way, by updating the pseudo-labels by clustering and by  discarding them from the whole training set during this clustering stage i.e. before the training stages. It can also be done in an online way, at the batch level, i.e. during the training stages. To perform offline filtering, DBSCAN is generally used to update the pseudo-labels and discard the erroneous ones. This clustering algorithm considers as outliers the samples belonging to clusters with a number of samples inferior to a number specified as a hyperparameter. \\
As for online filtering, state-of-the art approaches introduced advanced techniques such as Asymmetric Co-Teaching (ACT) (\cite{yang2019asymmetric})  or online label propagation (\cite{yang2019asymmetric, zheng2021online}). Since performance improvements have been shown when using DBSCAN (\cite{zheng2021online}), this suggests that online outlier filtering should be useful in addition to performing offline outlier filtering with DBSCAN. Therefore, Outlier Filtering should include offline and online outlier filtering. 
However, these existing strategies are method-specific and can easily increase the computation cost and resources needed for a Pseudo-Labeling approach where they are added: for example a teacher network (\cite{yang2019asymmetric}) or feature and label memory banks (\cite{zheng2021online}). 
Therefore, we propose a more simple and general online outlier filtering strategy based on an outlier detection statistical test: the Tukey Criterion (\cite{bliss1956rejection}). This criterion is applied on the loss values. Indeed, the intuition behind using the loss values for filtering, as for ACT (\cite{yang2019asymmetric}), is that uncommonly high values of the loss function are more likely to correspond to outliers. Therefore, computing the Tukey Criterion gives us a loss threshold above which the target domain samples are considered as outlier and thus are discarded from the batch.
It is also possible to propose  a lighter version of ACT, where the teacher is the model itself. Basically, we filter out the target samples corresponding to the top-$p$\% ($p \in \mathbb{N}$) highest loss values in every batch. In the rest of the paper, experiments are conducted with DBSCAN and the Tukey Criterion for the implementation of Outlier Filtering. The Tukey Criterion needs a confidence coefficient of rejecting the null hypothesis set to 0.05. 

\noindent \textbf{Bounded Loss.} The Bounded Loss aims at controlling the loss bound $M$. An easy way to control the loss bound would be to use a bounded loss for re-ID learning. However, in practice, the ``0-M" loss cannot be used for re-ID learning with gradient descent, because it is not differentiable. Indeed, other classical losses are used for re-ID such as the Cross-Entropy classification loss, the Triplet Loss or the Contrastive loss (\cite{ge2020self}). In order to avoid changing the loss functions of the UDA re-ID method, which are sometimes at the core of the UDA re-ID approach, we propose a more flexible strategy to controll their bound, that we call Thresholding/Normalization. For the Triplet Loss, that optimizes directly on the distances between features, Thresholding/Normalization normalizes the features, which consequently bound the distances on the unit norm ball, and thus the loss function. Moreover, for the other loss functions, Thresholding/Normalization thresholds the values of the loss above a defined threshold. The samples associated to the loss values above the threshold are discarded from the batch and therefore will not be back-propagated for parameter updates by gradient descent. In order not to introduce a new arbitrarily-set hyperparameter for this loss threshold, the threshold obtained by the Tukey Criterion used for Outlier Filtering, described in the previous paragraph, can be reused to choose the loss threshold.\\
While UDA re-ID classically uses unbounded losses, for the classification task, bounded classification losses robust to label noise have been designed. In particular, Ghosh et al. (\cite{ghosh2017robust}) show in their paper that the Mean Absolute Error (MAE) is an unbounded loss which improves empirically the classification performance when the labels are corrupted by different amount of noises. Moreover, it can be easily computed as the L1 penalization of the difference between 1 and the predicted probability for the ground-truth class by the model. Therefore, this could be another candidate to implement Bounded Loss. \\
In the rest of the paper, the Thresholding/Normalization strategy is used as Bounded Loss.

\noindent \textbf{Model Regularization.} Model Regularization aims at limiting the model complexity to prevent overfitting. In practice, Model Regularization can be performed with general regularization techniques, such as Weight Decay that penalizes the L2 norm of the model parameters. Or they can specifically be chosen and designed to be robust to noisy pseudo-labels, such as Mutual Mean Teaching (\cite{ge2020mutual}).
Model Regularization is to our knowledge followed by all Pseudo-Labeling UDA re-ID methods which always use weight decay. Moreover, specific Model Regularization implementations, designed to be robust to noisy labels are generally inherent to the framework design: changing them or performing an ablation of them would completely distort the method. For these reasons, in future experiments, we choose to keep as they are the Model Regularization of the Pseudo-Labeling baselines of interest.

\subsection{State-of-the-art baselines.}
\label{sec:baselines}

\begin{table*}
\centering
\caption{This table represents good practices already followed in the four original state-of-the-art frameworks of interest: UDAP (\cite{song2020unsupervised}), MMT (\cite{ge2020mutual}), SpCL (\cite{ge2020self}) and SpCL + HyPASS (\cite{dubourvieux2021improving}), compared to their respective version following all good practices (w/ all good practices). The missing good practice implementations are represented in bold and green. WD = Weight Decay ; MMD = Maximum Mean Discrepancy.}
\label{table:good_practice}
\resizebox{\linewidth}{!}{

\begin{tabular}{@{}cccccc@{}} 
\toprule
 Method   & Bounded loss          & Outlier Filtering               & Source-guided Learning & Domain Alignment      & Model Regularization      \\ 
\midrule
UDAP                                       & $\times$                       & DBSCAN  (\checkmark)                         & $\times$               & $\times$                & WD (\checkmark)                       \\ 
\hdashline
UDAP w/ all good practices                      & \textcolor{ForestGreen}{\textbf{Thresholding/Normalization} (\checkmark)}        & \makecell[c]{DBSCAN (\checkmark)\\ \textcolor{ForestGreen}{\textbf{ Tukey Online Filtering (\checkmark)}}} &   \textcolor{ForestGreen}{\textbf{\checkmark}}                    & \textcolor{ForestGreen}{\textbf{MMD} (\checkmark)}                   & WD  (\checkmark)                      \\ 
\midrule
MMT                                        & $\times$                     & $\times$                       & $\times$                & $\times$                & \makecell[c]{WD (\checkmark)\\ Mutual Mean Teaching (\checkmark)}  \\ 
\hdashline
MMT w/ all good practices                      & \textcolor{ForestGreen}{\textbf{Thresholding/Normalization} (\checkmark)}            &  \makecell[c]{ \textcolor{ForestGreen}{\textbf{DBSCAN (\checkmark)}}\\ \textcolor{ForestGreen}{\textbf{ Tukey Online Filtering (\checkmark)}}} &   \textcolor{ForestGreen}{\textbf{\checkmark}}                    & \textcolor{ForestGreen}{\textbf{MMD} (\checkmark)}            & \makecell[c]{WD (\checkmark)\\ Mutual Mean Teaching (\checkmark)}                        \\ 
\midrule
SpCL                                       & $\times$                       & \makecell[c]{DBSCAN (\checkmark) \\ Reliable Clusters (\checkmark)}       &         \checkmark              & $\times$                 & \makecell[c]{WD (\checkmark)\\ Hybrid Memory (\checkmark)}         \\ 
\hdashline
SpCL w/ all good practices                      & \textcolor{ForestGreen}{\textbf{Thresholding/Normalization} (\checkmark)}           & \makecell[c]{DBSCAN (\checkmark) \\ Reliable Clusters (\checkmark) \\ \textcolor{ForestGreen}{\textbf{Tukey Online Filtering (\checkmark)}}} &      \checkmark                 & \textcolor{ForestGreen}{\textbf{MMD} (\checkmark)}           & \makecell[c]{WD (\checkmark)\\ Hybrid Memory (\checkmark)}                        \\ 
\midrule
SpCL + HyPASS                                      & $\times$                       & \makecell[c]{DBSCAN (\checkmark) \\ Reliable Clusters (\checkmark) \\ Auto. Clustering (\checkmark)}      &         \checkmark              & MMD (\checkmark)               & \makecell[c]{WD (\checkmark)\\ Hybrid Memory (\checkmark)}         \\ 
\hdashline
SpCL + HyPASS  w/ all good practices                      & \textcolor{ForestGreen}{\textbf{Thresholding/Normalization} (\checkmark)}          & \makecell[c]{DBSCAN (\checkmark) \\ Reliable Clusters (\checkmark) \\ Auto. Clustering (\checkmark) \\ \textcolor{ForestGreen}{\textbf{Tukey Online Filtering  (\checkmark)}}} &      \checkmark                 & MMD (\checkmark)           & \makecell[c]{WD (\checkmark)\\ Hybrid Memory (\checkmark)}                       \\ 
\bottomrule
\end{tabular}
}
\end{table*}

We focus on four different state-of-the-art Pseudo-Labeling UDA re-ID baselines. As it will be detailed hereafter, our choice has been motivated by the fact that these baselines follow a varied inventory of good practices. Moreover, they correspond to recent UDA re-ID methods, with state-of-the art performances and an available code for reproducibility.\\

\noindent \textbf{UDAP (\cite{song2020unsupervised}).} UDAP is one of the first approaches using pseudo-labels for the UDA re-ID. In the learning process, the source data is only used to pretrain the feature encoder which initializes the first pseudo-labels. This approach uses DBSCAN as a clustering algorithm and therefore follows Outlier Filtering, more particularly offline Outlier Filtering, outside of the training loops. Moreover, UDAP minimizes the Triplet Loss, which is unbounded by definition. Finally, UDAP does not use any specific regularization on the model parameters apart from the classical weight decay.\\

\noindent \textbf{MMT (\cite{ge2020mutual}).} Like UDAP, MMT does not exploit the source data when training with the pseudo-labels. All the target samples are pseudo-labeled by k-means clustering algorithm and used for training: MMT does not perform Outlier Filtering at all. MMT optimizes both a Triplet Loss and a Cross-Entropy loss, with hard and soft pseudo-labels: these loss functions are also unbounded. Finally, MMT relies on the mutual learning paradigm and a mean teacher updated by an exponential moving average of the student model parameters. In semi-supervised learning, Mean Teachers (\cite{tarvainen2017mean}) as well as Mutual Learning (\cite{zhang2018deep}) are seen as consistency regularization by ensembling. Therefore, we can consider the Mutual Mean Teaching of MMT as a specific Model Regularization technique, used in addition to weight decay.\\

\noindent \textbf{SpCL (\cite{ge2020self}).} Unlike the UDAP and MMT approaches, SpCL leverages the source data during the training phases, in addition to the pseudo-labeled target data. However, SpCL does not perform Domain Alignment to reduce the domain discrepancy in the feature space. Moreover, a contrastive loss is optimized, which is an unbounded loss function. In addition to the Outlier Filtering performed by DBSCAN, SpCL further filters outliers with an additional cluster reliability criterion: Reliable Clusters. This criterion performs 2 clustering of the target dataset by DBSCAN, with 2 different density hyperparameter, and discard samples inconsistent between the 2 clusterings. Finally, SpCL uses a moving average of class-centroid and instance feature memory bank to compute the class centroids. As MMT, it can be viewed as consistency regularization by temporal ensembling of feature, and therefore as a Model Regularization module.\\

\noindent \textbf{SpCL + HyPASS (\cite{dubourvieux2021improving}).} HyPASS is a paradigm that can be added to any Pseudo-Labeling UDA re-ID approach that allows for automatic hyperparameter selection of the clustering algorithm. It relies on domain alignment of feature similarities. By adding HyPASS to SpCL (SpCL + HyPASS), the UDA re-ID accuracy are improved compared to the original SpCL. SpCL + HyPASS follows more good practices compared to SpCL. First, the automatic hyperparameter selection for clustering allows for a better selection of hyperparameters for the clustering stages, and therefore improves the quality of the pseudo-labels. Moreover, it performs Domain Alignment by penalizing the MMD loss between the source and target feature similarities, which contributes to a better selection of clustering hyperparameter with a source validation set. However, SpCL + HyPASS does not use a Bounded Loss. Moreoever, it will also be interesting to see if adding online Outlier Filtering could improve the UDA re-ID performance of SpCL + HyPASS.\\

In summary, the existing Pseudo-Labeling approaches for UDA re-ID do not follow all good practices (as summarized in Tab.~\ref{table:good_practice}). To experimentally validate these good practices derived from the theory, as well as their generalization to different UDA re-ID frameworks, we propose to follow the missing ones to these state-of-the-art approaches with using the implementation discussed in Sec.~\ref{sec:exp_gp}.

\subsection{Implementation details.}
\label{sec:exp_details}
We reused the codes\footnote{https://github.com/LcDog/DomainAdaptiveReID} \footnote{https://github.com/yxgeee/MMT} \footnote{https://github.com/yxgeee/SpCL} given by the baseline authors as well as the same implementation details (learning rate, architecture,...) in their respective paper (\cite{song2020unsupervised,ge2020mutual,ge2020self,dubourvieux2021improving}). For MMT, different architectures and hyperparameter values for the number of clusters for k-means are used in the paper. To allow fair comparison with other frameworks, we choose the ResNet-50 (\cite{he2016deep}) architecture and report the best performance of MMT among the number of clusters tested in the paper. For cross-dataset benchmarks that are not available in their paper, we use the number of ground-truth clusters of the target training set for the number of clusters. 
We use 4 x 24Go NVIDIA TITAN RTX GPU for all of our experiments.

\subsection{Cross-dataset benchmarks.} 
\label{sec:datasets}

\begin{table}[t]
\caption{Dataset composition}
\label{table:dataset}
\Scale[1]{
\begin{tabular}{@{}cccccc@{}} 
\toprule
Dataset   & \# train IDs & \# train images & \# test IDs & \# gallery images & \# query images  \\ \midrule
Market (\cite{zheng2015scalable})   & 751       & 12,936        & 750      & 16,364          & 3,368         \\
Duke (\cite{ristani2016performance})   & 702       & 16,522        & 702      & 16,364          & 2,228    \\
PersonX (\cite{sun2019dissecting})     & 410       & 9,840        & 856      & 17,661          & 30,816    \\
MSMT  (\cite{wei2018person})    & 1,041      & 32,621        & 3,060     & 82,161          & 11,659    \\ \hdashline
Vehicle-ID (\cite{liu2016deep})  & 13,164     & 113,346       & 800      & 7,332           & 6,532    \\
Veri (\cite{liu2016deep2})    & 575       & 37,746        & 200      & 49,325          & 1,678    \\ 
VehicleX (\cite{liu2016deep2})    & 1,362       & 192,150        & N.A.      & N.A.          & N.A.   \\ 
\bottomrule
\end{tabular}}
\end{table}

We study the effectiveness of good practices on the re-ID performance by computing and reporting the mean Average Precision (mAP in \%) and rank-1 (\%) on the target test sets for different cross-dataset adaptation tasks. It is important to have a variety of source and target datasets, since as the theory suggests, the bound depends on properties specific to the datasets (proportion of source/target samples in the dataset, domain discrepancy...).
\textit{Person re-ID} is evaluated on the large re-ID dataset MSMT17 (\cite{wei2018person}) (\textit{MSMT}): used as the target domain, it offers a challenging adaptation task due to its large number of images and identities in its gallery (cf. dataset statistics in Tab.~\ref{table:dataset}). We also use Market-1501 (\cite{zheng2015scalable}) (\textit{Market}) as the target domain, using the synthetic dataset PersonX as the source domain. \textit{PersonX} (\cite{sun2019dissecting}) is composed of synthetic images generated on Unity with different types of person appearances, camera views and occlusions. Then we also report classical benchmarks between Market and DukeMTMC-reID (\cite{ristani2016performance}) (\textit{Duke}). 
\textit{Vehicle re-ID} is less reported than Person re-ID for UDA re-ID benchmarking. However, we find it interesting to test our module on a different kind of object of interest and on a potentially different domain discrepancy. We use for this task  the \textit{Vehicle-ID} (\cite{liu2016deep}), Veri-776 (\cite{liu2016deep2}) (\textit{Veri}) datasets and the synthetic vehicle dataset \textit{VehicleX} (\cite{naphade20204th}).

\begin{center}
\begin{table*}[t!]
\caption{Comparison of original baselines with their corresponding version to which the missing good practices have been implemented (w/ all good practices) for different person cross-dataset benchmarks. mAP and rank-1 are reported in \%.}
\label{table:good_practice_person}
\resizebox{\linewidth}{!}{
\begin{tabular}{@{}ccccccccccccccl@{}} 
\toprule
\multirow{2}{*}{Method} &
  \multicolumn{2}{c}{Market$\rightarrow$MSMT} &
  \multicolumn{2}{c}{PersonX$\rightarrow$Market} &
  \multicolumn{2}{c}{PersonX$\rightarrow$MSMT} &
  \multicolumn{2}{c}{Market$\rightarrow$Duke} &
  \multicolumn{2}{c}{Duke$\rightarrow$Market} \\ \cmidrule{2-11}
 & mAP  & rank-1 & mAP  & rank-1 & mAP  & rank-1 & mAP  & rank-1 & mAP & rank-1 \\ \midrule
\multicolumn{1}{c}{UDAP (\cite{song2020unsupervised})} & 12.0 & 30.6     & 48.4 & 68.4     & 10.5 & 26.3     & 50.1 & 70.2     & 55.3  & 78.1     \\ 
\multicolumn{1}{c}{UDAP (\cite{song2020unsupervised}) w/ all good practices} &
  \textbf{20.9} &
  \textbf{47.0} &
  \textbf{67.1} &
  \textbf{70.8} &
  \textbf{14.9} &
  \textbf{36.1} &
  \textbf{63.1} &
  \textbf{77.3} &
  \textbf{69.4} &
  \textbf{86.5} \\ \midrule
\multicolumn{1}{c}{MMT (\cite{ge2020mutual})} &
  22.9 &
  49.2 &
  70.8 &
  66.8 &
  16.9 &
  38.5 &
  65.1 &
  78.0 &
  71.2 &
  87.7 \\ 
\multicolumn{1}{c}{MMT (\cite{ge2020mutual}) w/ all good practices} &
  \textbf{25.1} &
  \textbf{52.9} &
  \textbf{73.4} &
  \textbf{88.0} &
  \textbf{18.9} &
  \textbf{43.2} &
  \textbf{67.9} &
  \textbf{82.0} &
  \textbf{75.5} &
  \textbf{88.8} \\ \midrule
\multicolumn{1}{c}{SpCL (\cite{ge2020self})} & 25.7 & 53.4 & 72.2 & 86.1   & 22.1 & 47.7     & 68.3 & 82.5   & 76.1  & 89.8    \\ 
\multicolumn{1}{c}{SpCL (\cite{ge2020self}) w/ all good practices}                & \textbf{27.0} & \textbf{53.9}     & \textbf{74.1} & \textbf{88.6}     & \textbf{23.0} & \textbf{47.8}     & \textbf{70.6}   & \textbf{83.8}     & \textbf{78.0} & \textbf{91.4}\\ \midrule
\multicolumn{1}{c}{SpCL + HyPASS (\cite{dubourvieux2021improving})} & 27.4 & 55.0     & 77.9 & 91.5     & 23.7 & 48.6     & 71.1 & 84.5 & 78.9 & 92.1    \\ 
\multicolumn{1}{c}{SpCL + HyPASS (\cite{dubourvieux2021improving}) w/ all good practices}                       & \textbf{29.5} & \textbf{56.9}     & \textbf{79.4} & \textbf{93.1}     & \textbf{24.6} & \textbf{48.9}     & \textbf{72.1}   & \textbf{86.0}     & \textbf{80.2} & \textbf{93.4}\\    \bottomrule
\end{tabular}
}
\end{table*}
\end{center}

\begin{center}
\begin{table*}[t!]
\caption{Comparison of original baselines and their corresponding version to which the missing good practices have been implemented (w/ all good practices) for different vehicle cross-dataset benchmarks. mAP and rank-1 are reported in \%.}
\label{table:good_practice_vehicles}
\centering
\resizebox{\columnwidth}{!}{
\begin{tabular}{@{}ccccccccccccccl@{}} 
\toprule
\multirow{2}{*}{Method} &
  \multicolumn{2}{c}{VehicleID$\rightarrow$Veri} &
  \multicolumn{2}{c}{VehicleX$\rightarrow$Veri} \\ \cmidrule{2-6} & mAP & rank-1  & mAP & rank-1\\ \midrule
\multicolumn{1}{c}{UDAP (\cite{song2020unsupervised})} & 35.6 & 74.1     & 35.0  & 75.9     \\
\multicolumn{1}{c}{UDAP (\cite{song2020unsupervised}) w/ all good practices} &
  \textbf{37.1} &
  \textbf{75.2} &
  \textbf{37.2} &
  \textbf{77.1} \\ \midrule
\multicolumn{1}{c}{MMT (\cite{ge2020mutual})} &
  36.4 &
  74.2 &
  36.3 &
  75.8 \\ 
\multicolumn{1}{c}{MMT (\cite{ge2020mutual}) w/ all good practices} &
  \textbf{37.9} &
  \textbf{80.1} &
  \textbf{37.7} &
  \textbf{81.2} \\ \midrule
\multicolumn{1}{c}{SpCL (\cite{ge2020self})} & 37.6 & 79.7     & 37.4 & 81.0  \\ 
\multicolumn{1}{c}{SpCL (\cite{ge2020self}) w/ all good practices}  & \textbf{39.0}   & \textbf{83.4}     & \textbf{39.2} & \textbf{83.8}\\ \midrule
\multicolumn{1}{c}{SpCL + HyPASS (\cite{dubourvieux2021improving})} & 40.0   & 81.1     & 40.3 & 81.9  \\  
\multicolumn{1}{c}{SpCL + HyPASS (\cite{dubourvieux2021improving}) w/ all good practices}                      & \textbf{40.9}   & \textbf{85.7}     & \textbf{41.0} & \textbf{86.0}\\  \bottomrule

\end{tabular}
}
\end{table*}
\end{center}

\section{Experimental results}
\label{sec:results}

In Sec.~\ref{sec:good_exp}, we confirm by extensive experiments on various person and vehicle re-ID cross-datasets benchmarks, that implementing the missing good practices in these baselines can in fact improve their UDA re-ID performance. By ablative study in Sec.\ref{sec:ablative}, we analyze the contribution to the UDA re-ID performance of each followed good practice. Finally, in Sec.~\ref{sec:other_strategies}, we show by additional experiments, consistency of UDA re-ID performance improvement when good practices are implemented differently.

\subsection{Improving UDA re-ID frameworks by following good practices.}
\label{sec:good_exp}

In Tab.~\ref{table:good_practice_person} and Tab.~\ref{table:good_practice_vehicles}, we reported the mAP in \% obtained after adding the missing good practices in different baselines (see Tab.~\ref{table:good_practice}) resp. for person cross-dataset and vehicle cross-dataset benchmarks. We notice that for all frameworks UDAP, MMT, SpCL and SpCL + HyPASS, following good practices consistently improves performance on all person and vehicle adaptation tasks. More specifically, we notice that this improvement depends on how many elements of good practices are added in relation to the original baseline. The more missing good practices a method has, the more it is likely to benefit from following all good practices. For example, UDAP, which follows only 2 good practices, gains +8.9 p.p. mAP on Market$\rightarrow$MSMT, while SpCL + HyPASS, with the greatest number of good practices already followed, gains +2.1 p.p. mAP by following the missing good practices. 
Consistently, on Vehicle-ID$\rightarrow$Veri, UDAP gains +1.5 p.p. mAP, while SpCL + HyPASS gains +0.9 p.p. mAP by enforcing the missing good practices. Overall, enforcing good practices systematically improve performance whatever the number of missing good practices a framework follows. The next section aims at analyzing more thoroughly these performance gains with an ablative study.

\subsection{Ablation study on good practices.}
\label{sec:ablative}

As seen in Sec.~\ref{sec:good_exp}, following all good practices in a Pseudo-Labeling UDA re-ID framework improves cross-dataset performance. To quantify the contribution of each good practices individually, as well as to understand their interactions, we carried out on 2 cross-dataset benchmarks (PersonX$\rightarrow$Market and Market$\rightarrow$MSMT) an ablative study for each framework: UDAP, MMT, SpCL and SpCL + HyPASS resp. in Tab.~\ref{table:ablative_udap}, Tab.~\ref{table:ablative_mmt}, Tab.~\ref{table:ablative_spcl} and Tab.~\ref{table:ablative_spcl_HyPASS}.\\

\begin{table*}[t!]
\caption{Ablative study comparing UDAP to a version of UDAP where each of good practices has been followed individually. The missing and newly-followed good practices are represented in bold and green. mAP are reported in \% for two cross-dataset UDA re-ID tasks: PersonX$\rightarrow$Market and Market$\rightarrow$MSMT.\label{table:ablative_udap}}
\resizebox{\linewidth}{!}{
\begin{tabular}{@{}ccccccc@{}}
\toprule Variants &
 Bounded Loss &
  Outlier Filtering &
  Source-guided Learning &
  Domain Alignment &
  PersonX$\rightarrow$Market &
  Market$\rightarrow$MSMT \\ \midrule
 UDAP                                       & $\times$                     & DBSCAN (\checkmark)                          & $\times$             & $\times$               &
  48.4 &
  12.0 \\ \hdashline
1 &
  \textcolor{ForestGreen}{\textbf{Thresholding/Normalization} (\checkmark)} &
  DBSCAN (\checkmark) &
  $\times$ &
  $\times$  &
  50.3 &
  14.3 \\ \hdashline
2 &
  $\times$ &
  \makecell[c]{DBSCAN (\checkmark)\\ \textcolor{ForestGreen}{\textbf{ Tukey Online Filtering (\checkmark)}}} &
  $\times$ &
  $\times$ &
  59.7 &
  16.9 \\ \hdashline
3 &
  $\times$ &
  DBSCAN (\checkmark) &
  \textcolor{ForestGreen}{\textbf{\checkmark}} &
  $\times$ &
  53.1 &
  14.9 \\ \hdashline
4 &
  $\times$ &
  DBSCAN (\checkmark) &
  \textcolor{ForestGreen}{\textbf{\checkmark}} &
  \textcolor{ForestGreen}{\textbf{MMD} (\checkmark)}  &
  60.0 &
  16.8 \\ \hdashline
UDAP w/ all good practices
& \textcolor{ForestGreen}{\textbf{Thresholding/Normalization} (\checkmark)}          & \makecell[c]{DBSCAN (\checkmark)\\ \textcolor{ForestGreen}{\textbf{ Tukey Online Filtering (\checkmark)}}} &   \textcolor{ForestGreen}{\textbf{\checkmark}}                    & \textcolor{ForestGreen}{\textbf{MMD} (\checkmark)}    &
  \textbf{67.1} &
  \textbf{20.9} \\ \bottomrule
\end{tabular}
}
\end{table*}

\begin{table*}[t!]
\caption{Ablative study comparing MMT to a version of MMT where each of good practices has been followed individually. The missing and newly-followed good practices are represented in bold and green. mAP are reported in \% for two cross-dataset UDA re-ID tasks: PersonX$\rightarrow$Market and Market$\rightarrow$MSMT.}
\label{table:ablative_mmt}
\resizebox{\linewidth}{!}{
\begin{tabular}{@{}ccccccc@{}}
\toprule
 Variants &
  Bounded Loss &
  Outlier Filtering &
  Source-guided Learning &
  Domain Alignment &
  PersonX$\rightarrow$Market &
  Market$\rightarrow$MSMT \\ \midrule
MMT                                   & $\times$                     & $\times$                     & $\times$              & $\times$              &  70.8 & 22.9  \\ \hdashline
1 &
  \textcolor{ForestGreen}{\textbf{Thresholding/Normalization} (\checkmark)} &
  $\times$ &
  $\times$ &
  $\times$ &
  71.7 &
  24.5 \\ \hdashline
2 &
  $\times$ &
   \makecell[c]{\textcolor{ForestGreen}{\textbf{DBSCAN (\checkmark)}}\\ \textcolor{ForestGreen}{\textbf{ Tukey Online Filtering (\checkmark)}}} &
  $\times$ &
  $\times$ &
  
  72.2 &
  24.3 \\ \hdashline
3 &
  $\times$ &
  $\times$ &
  \textcolor{ForestGreen}{\textbf{\checkmark}} &
  $\times$ &
  
  72.9 &
  24.1 \\ \hdashline
4 &
  $\times$ &
  $\times$ &
  \textcolor{ForestGreen}{\textbf{\checkmark}} &
  \textcolor{ForestGreen}{\textbf{MMD} (\checkmark)} &
  
  73.0 &
  24.7 \\ \hdashline
MMT w/ all good practices                   & \textcolor{ForestGreen}{\textbf{Thresholding/Normalization} (\checkmark)}          & \makecell[c]{\textcolor{ForestGreen}{\textbf{DBSCAN (\checkmark)}}\\ \textcolor{ForestGreen}{\textbf{ Tukey Online Filtering (\checkmark)}}} &   \textcolor{ForestGreen}{\textbf{\checkmark}}                    & \textcolor{ForestGreen}{\textbf{MMD} (\checkmark)}            & 
  \textbf{73.4} &
  \textbf{25.1} \\ \bottomrule
\end{tabular}}
\end{table*}

\begin{table*}[t!]
\centering
\caption{Ablative study comparing SpCL to a version of SpCL where each of good practices has been followed individually. The missing and newly-followed good practices are represented in bold and green. mAP are reported in \% for two cross-dataset UDA re-ID task: PersonX$\rightarrow$Market and Market$\rightarrow$MSMT. \label{table:ablative_spcl}}
\resizebox{\linewidth}{!}{
\begin{tabular}{@{}ccccccc@{}}
\toprule
 Variants
 &
  Bounded Loss &
  Outlier Filtering &
  Source-guided Learning &
  Domain Alignment &
  PersonX$\rightarrow$Market &
  Market$\rightarrow$MSMT \\ \midrule
SpCL & $\times$                     & \makecell[c]{DBSCAN (\checkmark)\\ Reliable Clusters (\checkmark)}       &         $\times$              & $\times$              & 
  72.2 &
  25.7 \\ \hdashline
1 &
  \textcolor{ForestGreen}{\textbf{Thresholding/Normalization} (\checkmark)} &
  \makecell[c]{DBSCAN (\checkmark) \\ Reliable Clusters (\checkmark)}  &
  \checkmark &
  $\times$ &
  
  72.9 &
  26.8 \\ \hdashline
2 &
  $\times$ &
  \makecell[c]{DBSCAN (\checkmark) \\ Reliable Clusters (\checkmark) \\ \textcolor{ForestGreen}{\textbf{ Tukey Online Filtering (\checkmark)}}} &
  \checkmark &
  $\times$ &
  
  73.3 &
  26.6 \\ \hdashline
4 &
  $\times$ &
  \makecell[c]{DBSCAN (\checkmark) \\ Reliable Clusters (\checkmark)}  &
  \checkmark &
  \textcolor{ForestGreen}{\textbf{MMD} (\checkmark)} &
  
  73.2 &
  26.5 \\ \hdashline
SpCL w/ all good practices                      & \textcolor{ForestGreen}{\textbf{Thresholding/Normalization} (\checkmark)}          & \makecell[c]{DBSCAN (\checkmark) \\ Reliable Clusters (\checkmark) \\ \textcolor{ForestGreen}{\textbf{Tukey Online Filtering (\checkmark)}}} &      \checkmark                 & \textcolor{ForestGreen}{\textbf{MMD} (\checkmark)}           & 
  \textbf{74.1} &
  \textbf{27.0} \\ \bottomrule
\end{tabular}}
\end{table*}

\begin{table*}[h!]
\centering
\caption{Ablative study comparing SpCL + HyPASS to a version of SpCL + HyPASS wherewhere each of good practices has been followed individually. The missing and newly-followed good practices are represented in bold and green. mAP are reported in \% for two cross-dataset UDA re-ID task: PersonX$\rightarrow$Market and Market$\rightarrow$MSMT.}
\label{table:ablative_spcl_HyPASS}
\resizebox{\linewidth}{!}{
\begin{tabular}{@{}ccccccc@{}}
\toprule
Variants &
  Bounded Loss &
  Outlier Filtering &
  Source-guided Learning &
  Domain Alignment &
  PersonX$\rightarrow$Market &
  Market$\rightarrow$MSMT \\ \midrule
SpCL + HyPASS & $\times$                     & \makecell[c]{DBSCAN (\checkmark) \\ Reliable Clusters (\checkmark) \\ Auto. Clustering (\checkmark)}       &         $\times$               & MMD  (\checkmark)            & 
  77.9 &
  27.4 \\ \hdashline
1 &
  \textcolor{ForestGreen}{\textbf{Thresholding/Normalization} (\checkmark)} &
  \makecell[c]{DBSCAN (\checkmark) \\ Reliable Clusters (\checkmark) \\ Auto. Clustering (\checkmark)}  &
  \checkmark &
  MMD (\checkmark) &
  
  78.5 &
  28.3 \\ \hdashline
2 &
  $\times$  &
  \makecell[c]{DBSCAN (\checkmark) \\ Reliable Clusters (\checkmark) \\ Auto. Clustering (\checkmark) \\ \textcolor{ForestGreen}{\textbf{ Tukey Online Filtering (\checkmark)}}} &
  \checkmark &
  MMD (\checkmark) &
  
  78.6 &
  28.2 \\ \hdashline
SpCL + HyPASS w/ all good practices                      & Thresholding/Normalization (\checkmark)          & \makecell[c]{DBSCAN (\checkmark) \\ Reliable Clusters (\checkmark) \\ Auto. Clustering (\checkmark) \\ \textcolor{ForestGreen}{\textbf{ Tukey Online Filtering (\checkmark)}}} &      \checkmark                 & \textcolor{ForestGreen}{\textbf{MMD} (\checkmark)}           & 
  \textbf{79.4} &
  \textbf{29.5} \\ \bottomrule
\end{tabular}}
\end{table*}

\noindent \textbf{Bounded Loss.} Comparing the original frameworks with their variant 1, which bounds the re-ID loss function by Thresholding/Normalization, we note that the Bounded Loss good practice allows an improvement of the re-ID performance for all frameworks and the two cross-dataset UDA re-ID tasks. For PersonX$\rightarrow$Market, variant 1, with the Bounded Loss good practice followed, improves the original framework resp. by +1.9 p.p., +0.9 p.p., +0.7 p.p. and +0.6 p.p. mAP for UDAP, MMT, SpCL and SpCL + HyPASS. Consistently, for Market$\rightarrow$MSMT, variant 1, improves the original framework resp. by +2.3 p.p., +1.6 p.p., +1.1 p.p. and +0.9 p.p. mAP for UDAP, MMT, SpCL and SpCL + HyPASS. This is in line with our analysis of the learning bound, from which we deduce the Bounded Loss as a good practice improving the UDA re-ID performance. Moreover, we also notice that the less a framework follows Outlier Filtering and Model Regularization good practices, the more performance enhancement they benefit from Bounded Loss good practice: the mAP increase is higher for UDAP compared to MMT, higher for MMT compared to SpCL and higher for SpCL than for SpCL + HyPASS. We reckon it is in line with our theoretical analysis: it should be caused by the multiplicative interaction in the learning bound between the loss bound $M$ and the complexity term that is further reduced with better Model Regularization for MMT and SpCL, as well as the multiplicative interaction between $M$ and the Domain Discrepancy term $\mathcal{D}\mathcal{D}$ which is further reduced by Domain Alignment for SpCL + HyPASS.\\

\noindent \textbf{Outlier Filtering.} Variant 2, adding Outlier Filtering, improves performance over the original frameworks. For PersonX$\rightarrow$Market, Outlier Filtering improves the original framework by resp. +11.3 p.p., +1.1 p.p., +1.4 p.p and +0.7 p.p mAP for UDAP, MMT, SpCL and SpCL + HyPASS. Consistently, for Market$\rightarrow$MSMT, variant 2, improves the original framework resp. by +4.9 p.p., +1.4 p.p., +0.9 p.p. and +0.8 p.p. mAP for UDAP, MMT, SpCL and SpCL + HyPASS. 
Therefore, it seems to confirm that Outlier Filtering, and more generally reducing the noise probabilities in the learning bound, is a good practice for Pseudo-Labeling UDA re-ID. What's more, we also notice that this enhancement is more significant for UDAP and less for frameworks with better Model Regularization (MMT, SpCL, SpCL + HyPASS). Again, we guess that this result accounts for the multiplicative interaction in the learning bound between the Noise term $\mathcal{N}$ and the Complexity term $\mathcal{C}$ reduced by Model Regularization.\\

\noindent \textbf{Source-guided Learning with Domain Alignment.} By using the source samples to learn re-ID (variant 3 of UDAP and MMT) in Tab.~\ref{table:ablative_udap} and Tab.~\ref{table:ablative_mmt}, performance get improved over the original frameworks. For PersonX$\rightarrow$Market, Source-guided Learning improves the original framework resp. by +4.7 p.p and +2.1 p.p. mAP for UDAP and MMT. Consistently, for Market$\rightarrow$MSMT, Source-guided Learning improves the original framework resp. by +2.9 p.p. and +1.2 p.p. mAP for UDAP and MMT. 
This performance improvement is consistent with our learning bound analysis, which states that using the source samples should help to improve the UDA re-ID performance. Adding Domain Alignment (variant 4 for UDAP, MMT in Tab.~\ref{table:ablative_udap} and Tab.~\ref{table:ablative_mmt}, and SpCL in Tab~\ref{table:ablative_spcl}) further improves performance. For PersonX$\rightarrow$Market, Source-guided Learning with Domain Alignment improves the original framework resp. by +11.6 p.p., +2.2 p.p. and +1.0 p.p mAP for UDAP, MMT and SpCL. Consistently, for Market$\rightarrow$MSMT, adding Domain Alignment increases the performance resp. by +4.8 p.p., +1.8 p.p. and +0.8 p.p. mAP for UDAP, MMT and SpCL . It is therefore more effective to reduce the domain gap with Domain Alignment when using the source samples to learn re-ID.\\

\noindent \textbf{Model Regularization.} Model Regularization is one of good practices derived from our theoretical analysis. As discussed previously in Sec.~\ref{sec:bound}, we did not perform an ablation of Model Regularization from the Pseudo-Labeling approaches since their regularization techniques generally define the core of these approaches. Yet, to experimentally confirm the role of Model Regularization as a good practice, it is possible to compare UDAP w/ all good practices to MMT w/ all good practices. Indeed, MMT w/ all good practices corresponds to UDAP w/ all good practices to which we add the Mutual Mean Teaching model regularization. Therefore, MMT w/ all good practices improves the UDA re-ID performance resp. by +6.3 p.p. and +4.2 p.p. mAP for PersonX$\rightarrow$Market and Market$\rightarrow$MSMT. Model Regularization, specifically designed for preventing noise overfitting like MMT, seems therefore to be a key good practice to significantly boost the UDA re-ID performance.

\subsection{Experiments with other implementations of good practices.}

This section adds additional experiments to show that the derived good practices are still consistent when followed differently, particularly for Outlier Filtering and Domain Alignment for which other strategies exist.

\label{sec:other_strategies}

\begin{table}[h!]
\centering
\caption{Impact on the UDA re-ID performance of MMT when using different Outlier Filtering strategies. mAP are reported in \% for the cross-dataset task PersonX$\rightarrow$Market. The percentage of outlier in a batch after filtering, averaged over all the training iterations, are also reported in \%.}
\label{table:exp_outlier}
\Scale[0.8]{
\begin{tabular}{@{}ccc@{}}
\toprule
Outlier Filtering & PersonX$\rightarrow$Market  & Average of outliers per batch \\ \midrule
$\times$                 & 70.8 & 27.2             \\
DBSCAN                 & 71.1 & 19.3             \\
Tukey                 & 71.8 & 10.1             \\
DBSCAN + Tukey                 & 72.2 & 7.2             \\
top-5\%           & 71.1 & 18.4            \\
top-10\%          & 71.5 & 10.2            \\
top-20\%          & 70.9 & 3.9              \\
DBSCAN + top-10\%             & 72.0 & 8.0              \\ \bottomrule
\end{tabular}}
\end{table}

\noindent \textbf{Changing the Outlier Filtering strategy.} Here we change the way Outlier Filtering is performed in the framework. Experiments are conducted for MMT which does not perform any Outlier Filtering. Different Outlier Filtering strategies are evaluated: DBSCAN, Tukey, DBSCAN + Tukey, and top-5/10/20\% introduced in Sec.~\ref{sec:exp_gp}. The re-ID cross-dataset performance (mAP) is reported in Tab.~\ref{table:exp_outlier}, as well as the average percentage of outliers in a batch during all the training. This quantity is computed by counting the pairs of data with same pseudo-label yet different ground-truth labels as well as those with different pseudo-labels yet same ground-truth labels.
First, we notice that every Outlier Filtering strategy improves the original framework. The performance improvement ranges from +0.1 p.p. mAP with top-20\% to +1.4 p.p. mAP with DBSCAN + Tukey. These experiments seem to confirm that there are different ways of enforcing the Outlier Filtering good practice in a Pseudo-Labeling UDA re-ID framework.\\
Moreover, by analyzing the average percentage of outliers per batch, we can conclude that every Outlier Filtering strategy effectively reduces the number of outliers, as expected. Even if in general, a lower percentage of outliers may be correlated to a better final re-ID performance, the top-20\% strategy seems to be an exception. Indeed, the top-20\% strategy filters out more outliers than other strategies but does not offer the best re-ID performance in the end. We suppose that Outlier Filtering strategies may discard valuable training samples, particularly hard positive/negative samples, which are more likely to be selected as outlier whereas they represent valuable information to learn ID discriminative representation.

\begin{table}[h!]
\centering
\caption{Impact on the UDA re-ID performance when using different Domain Alignment strategies. mAP are reported in \% for the cross-dataset task PersonX$\rightarrow$Market on UDAP.}
\label{table:exp_alignment}
\resizebox{0.9\columnwidth}{!}{
\begin{tabular}{@{}cc@{}}
\toprule
Domain Alignment & PersonX$\rightarrow$Market  \\ \midrule
$\times$            & 48.4 \\
Domain Adversarial Neural Network (\cite{ganin2016domain})  & 60.7 \\
Optimal Transport (Sinkhorn (\cite{cuturi2013sinkhorn}) ) & 61.1 \\
MMD          & 60.0 \\ \bottomrule
\end{tabular}}
\end{table}

\noindent \textbf{Changing the Domain Alignment criterion.} We also conducted more experiments with UDAP for PersonX$\rightarrow$Market where the Domain Alignment criterion is changed when performing Source-guided Learning with Domain Alignment, with some implementations introduced in Sec.~\ref{sec:exp_gp}. In Tab.~\ref{table:exp_alignment}, for PersonX$\rightarrow$Market, whatever the Domain Alignment strategy, we notice performance improvements of UDAP ranging from +11.6 p.p. mAP for MMD to +12.7 p.p. mAP for Optimal Transport. Again, experiments show flexibility on the way used to enforce the Domain Alignment good practice.

\noindent \textbf{Changing the Bounded Loss.}

In the previous experiments, to follow the Bounded Loss good practice, we chose to bound the re-ID implemented in the framework using the threshold computed with the Tukey Criterion applied on the loss values. \\
In these experiments, the Triplet Loss of UDAP is replaced by the MAE loss introduced in Sec.~\ref{sec:exp_gp}. Experiments are conducted for PersonX$\rightarrow$Market. In Tab.~\ref{table:exp_loss}, using the MAE Loss to implement the Bounded Loss good practice, the cross-domain re-ID performance is improved by +2.1 p.p. mAP. As it has been shown for the Thresholding/Normalization, following the Bounded Loss good practice, with the MAE loss, indeed improves the cross-domain re-ID performance. This indicates that the Bounded Loss is a general good practice that can be implemented with different strategies to follow it.

\begin{table}[h!]
\centering
\caption{Impact on the UDA re-ID performance when using different Bounded Loss strategies. mAP are reported in \% for the cross-dataset task PersonX$\rightarrow$Market on UDAP.}
\label{table:exp_loss}
\resizebox{0.8\columnwidth}{!}{
\begin{tabular}{@{}cc@{}}
\toprule
Bounded Loss & PersonX$\rightarrow$Market  \\ \midrule
$\times$            & 48.4 \\
Thresholding/Normalization  & 50.3 \\
Mean Absolute Error (MAE) (\cite{ghosh2017robust}) & 50.5 \\ \bottomrule
\end{tabular}}
\end{table}

\section{Conclusion}

In this work, we have proposed a general theoretical view to Pseudo-Labeling approaches for UDA re-ID. Throughout a learning upper-bound  on the target re-ID performance, we have been able to get more insight, as well as establishing a theoretical framework, for the wide variety of state-of-the-art methods which aim at improving Pseudo-Labeling UDA re-ID. In addition, we were able to derive from the theory a set of general good practices for Pseudo-Labeling UDA re-ID. Noting that existing UDA re-ID frameworks generally do not apply all of the derived good practices, we have showed by experiments that the UDA re-ID can further be improved for these approaches by following the missing good practices. These good practices being general, we have also showed by experiment that they can be implemented in various ways, and consistently improve the performance of various state-of-the art methods, on several person and vehicle cross-domain benchmarks. We hope this work will foster research to bring more understanding to UDA re-ID. For future work, it could be interesting to take inspiration from this one, focusing on UDA re-ID, to bring insight to other UDA problems where Pseudo-Labeling approaches prevail. 

\section*{Acknowledgments}
This publication was made possible by
the use of the FactoryIA supercomputer, financially supported
by the Ile-de-France Regional Council.

\bibliographystyle{ieeetr}
\bibliography{main.bib}

\end{document}